\documentclass{tlp}

\usepackage{times}
\usepackage{alltt}
\usepackage{url}
\usepackage{xspace}
\def\naf{{\neg \:}}  

\newcommand{\nop}[1]{}

\usepackage{latexsym} 
\usepackage{epsfig}
\usepackage{amssymb}
\usepackage{amsmath}
\usepackage{url}
\newcommand{\lpw}{\dl[{w}]}

\newtheorem{theorem}{Theorem}[section]

\newtheorem{ex}[theorem]{Example}
\newenvironment{example}{\begin{ex} \rm}{\end{ex}}

\newtheorem{definition}[theorem]{Definition}
\newtheorem{proposition}[theorem]{Proposition}

\newcommand{\dlv}{{\bf\small{DLV}}}

\newtheorem{lemma}{Lemma}

\def\bldl{\smallskip\[\bf\begin{array}{ll}}
\def\cldl{\vspace{-0.4cm}\[\bf\begin{array}{ll}}
\def\eldl{\end{array}\]\rm}

\def\lrule#1#2{ #1 \derives #2 \ . }

\newcommand{\pr}[1]{\stepcounter{#1}(\arabic{#1})}

\def\qed{\hspace*{\fill}{\rule[-0.5mm]{1.5mm}{3mm}}}
\newcommand{\dl}[1][ ]{\ensuremath{\texttt{\sc LP}^{#1}}\xspace}

\newenvironment{dlvcode}
  {\begin{displaymath}\begin{array}{l}\tt}
  {\end{array}\end{displaymath}}

\newcommand{\derives}{:\!\!-}
\newcommand{\weakderives}{:\sim}
\newcommand{\tuple}[1]{\langle#1\rangle}

\newcommand{\set}[1]{\{#1\}}

\newcommand{\Pol}{{\rm P}}

\newcommand{\NP}{{\rm NP}}

\newcommand{\CONP}{\mbox{\rm co-}\NP }
\newcommand{\DP}[1]{{\rm D_{#1}^P}}
\newcommand{\SigmaP}[1]{{\Sigma}_{#1}^{P}}
\newcommand{\PiP}[1]{{\Pi}_{#1}^{P}}

\newcommand{\DeltaP}[1]{{\Delta}_{#1}^{P} }
\newcommand{\FPNP}{{\rm FP}^{\rm NP}}

\def\bbegin{{\bf begin}}
\def\bend{{\bf end}}
\def\breturn{{\bf return}}
\def\bvar{{\bf var}}
\def\bfor{{\bf for}}
\def\bto{{\bf to}}
\def\bdo{{\bf do}}
\def\bendfor{{\bf end\_for}}
\def\bif{{\bf if}}
\def\bthen{{\bf then}}
\def\belse{{\bf else}}
\def\bFunction{{\bf Function}}
\def\BP{B_{\gp}}

\def\gq{\ensuremath{\cal Q}}
\def\HP{\ensuremath{\mathcal{H}_\gp}}
\def\HPs{\ensuremath{\mathcal{H}_{\gp_s}}}

\newcommand{\gpst}{\gp\ = \tuple{H,P,O,\gamma}}
\newcommand{\lpwp}{\lpw({\cal P})}

\newcommand{\PAP}{\mbox{\it P\hspace*{-0.150em}A\hspace*{-0.14em}P}}

\newcommand{\OPT}{\mbox{\it O\hspace*{-0.075em}p\hspace*{-0.05em}t}}
\newcommand{\ADM}{\mbox{\it A\hspace*{-0.075em}d\hspace*{-0.05em}m}}

\def\gpd{P}

\def\gp{\ensuremath{{\cal P}} }

\def\BP {B_{\gpd}}

\newcommand{\SM}{{\it SM}}
\newcommand{\facts}{{\it facts}}
\newcommand{\true}{{\it true}}
\newcommand{\false}{{\it false}}
\newcommand{\incons}{{\it inconsistent}}
\newcommand{\contr}{{\it contr}}
\newcommand{\at}{{\it at}}

\newcommand{\visited}{{\it visited}}

\newcommand{\diff}{{\it diff}}
\newcommand{\ok}{{\it ok}}
\newcommand{\tmin}{t_{\it min}}

\newcommand{\p}{{\it P}}

\newcommand{\var}{{\it var}}

\title[Abductive Logic Programs with Penalization]
        {Abductive Logic Programs with Penalization: \\
Semantics, Complexity and Implementation}

\author[S.~Perri, F.~Scarcello, N.~Leone]
{
 SIMONA PERRI\\
 Department of Mathematics, University of Calabria\\
 87030 Rende (CS), Italy\\
 \email{perri@mat.unical.it} \\
 \and FRANCESCO SCARCELLO\\
 DEIS, University of Calabria\\
 87030 Rende (CS), Italy\\
 \email{scarcello@deis.unical.it}\\
 \and NICOLA LEONE\\
 Department of Mathematics, University of Calabria\\
 87030 Rende (CS), Italy\\
 \email{leone@mat.unical.it} \\
}

\submitted{28 Januari 2003} \revised{16 October 2003} \accepted{21
October 2003}
\begin{document}
\maketitle

\begin{abstract}
Abduction, first proposed in the setting of classical logics, has
been studied with growing interest in the logic programming area
during the last years.

In this paper we study {\em abduction with penalization}
in the logic programming framework.
This form of abductive reasoning, which has not been
previously analyzed in logic programming, turns out to represent
several relevant problems, including optimization problems, very
naturally. We define a formal model for abduction with
penalization over logic programs, which extends the abductive
framework proposed by Kakas and Mancarella.
We address knowledge representation issues,
encoding a number of problems in our abductive framework.
In particular, we consider some relevant problems,
taken from different domains, ranging from optimization theory to diagnosis
and planning; their encodings turn out to be simple and elegant in our
formalism.
We thoroughly analyze the computational complexity
of the main problems arising in the context of abduction with
penalization from logic programs.
\nop{An interesting result in this course is that ``negation comes
for free.'' Indeed, the addition of negation
does not cause any further increase to the complexity of the
abductive reasoning tasks (which remains the same as for not-free
programs).}
Finally, we implement a system supporting the proposed abductive framework
on top of the \dlv\ engine.
To this end, we design a translation from
abduction problems with penalties into logic programs with weak
constraints. We prove that this approach is sound and complete.
\end{abstract}


\newpage

\section{Introduction}\label{sec:intro}

Abduction is an important form of reasoning, first studied
in depth by Peirce \shortcite{peir-55}.
 Given the observation of some facts, abduction aims at
concluding the presence of other facts, from which, together with an
underlying theory, the observed facts can be explained, i.e.,
deductively derived.
Thus, roughly speaking, abduction amounts to an inverse of modus
ponens.

For example, medical diagnosis is a typical abductive
reasoning process: from the symptoms and the medical knowledge, a
diagnosis about a possible disease is abduced. Notice that this form
of reasoning is not sound (a diagnosis may turn out to be wrong), and that
in general several abductive explanations (i.e., diagnoses) for
the observed symptoms may be possible.

It has been recognized that abduction is an important principle of
common-sense reasoning, and that abduction has fruitful
applications in a number of areas such diverse as model-based
diagnosis \cite{pool-89a}, speech recognition \cite{hobb-etal-93},
model checking \cite{bucc-etal-99a-ai},
maintenance of database views \cite{kaka-manc-90}, and vision
\cite{char-mcde-85}.

Most research on abduction concerned abduction from
classical logic theories.
However, there are several application domains where
the use of logic programming to perform abductive reasoning
seems more appropriate and natural \cite{eite-etal-97k}.

For instance, consider the following scenario. Assume that it is Sunday and is
known that Fabrizio plays soccer on Sundays if it's not raining. This may be
represented by the following theory $T$:
$$  play\_soccer \leftarrow is\_sunday \wedge not \ rains
\qquad\qquad is\_sunday\leftarrow $$ Now you observe that Fabrizio
is not out playing soccer (rather, he is writing a paper).
Intuitively, from this observation we conclude that it rains (i.e,
we abduce $rains$), for otherwise Fabrizio would be out playing
soccer. Nevertheless, under classical inference, the fact $rains$
is not an explanation of $not \ play\_soccer$, as $T \cup
\{rains\} \not \models not \ play\_soccer$ (neither can one find
any explanation). On the contrary, if we adopt the semantics of
logic programming (interpreting $not \ $ as the nonmonotonic
negation operator), then, according with the intuition, we obtain
that $rains$ is an explanation of $not \ play\_soccer$, as it is
entailed by  $T \cup \{rains\}$.

In the context of logic programming, abduction has been first
proposed by Kakas and Mancarella \shortcite{kaka-manc-90b} and,
during the recent years, the interest in this subject has been
growing rapidly
\cite{cons-etal-91,kono-92,kaka-etal-92,dung-91,dene-desc-95,saka-inou-00,bren-98,kaka-etal-00,dene-kaka-02,lin-you-02}.
This is also due to some advantages in dealing with incomplete
information that this kind of reasoning has over deduction
\cite{dene-desc-95,bara-gelf-94}.

Unlike most of these previous works on abduction in the logic
programming framework, in this paper we study {\em abduction with
penalization} from logic programs. This form of abductive
reasoning, well studied in the setting of classical logics
\cite{eite-gott-95}, has not been previously analyzed in logic
programming.

Note that dealing with weights or penalties has been recognized as a
very important feature of knowledge representation systems.
In fact, even at the very recent Workshop on Nonmonotonic Reasoning,
Answer Set Programming and Constraints (Dagstuhl, Germany, 2002),
many talks and system demonstrations pointed out that a lot of
problems arising in real applications requires the ability to
discriminate over different candidate solutions, by means of some suitable
preference relationship.
Note that this is not just an esthetic issue, for
representing such problems in a more natural and
declarative way. Rather, a proper use of preferences may have a
dramatic impact even on the efficiency of solving these problems.
\vspace{0.2cm}

In this paper, we define a formal model for abduction with
penalization from logic programs, which extends the abductive
framework proposed by Kakas and
Mancarella~\shortcite{kaka-manc-90b}. Roughly, a problem of
abduction with penalization $\gp$ consists of a logic program $
P$, a set of hypotheses, a set of observations, and a function
that assigns a penalty to each hypothesis. An admissible solution
is a set of hypotheses such that all observations can be derived
from $P$ assuming that these hypotheses are true. Each solution is
weighted by the sum of the penalties associated with its
hypotheses. The optimal solutions are those with the minimum
weight, which are considered more likely to occur, and thus are
preferred over other solutions with higher penalties.

We face knowledge representation issues, by showing how abduction with
penalization from logic programming can be used for encoding easily
and in a natural way relevant problems belonging to different domains.
In particular, we consider the classical
{\em Travelling Salesman Problem} from optimization theory,
(a new version of) the {\em Strategic Companies Problem},
the planning problem {\em Blocks World}, from artificial intelligence.
It is worthwhile noting that these problems
cannot be encoded at all in (function-free) normal logic programming,
even under the powerful stable model semantics.
\vspace{0.2cm}

We analyze the computational
complexity of the main problems arising in this framework, namely,
given a problem $\gp$ of abduction with penalization over logic programs,
\begin{itemize}

\item decide whether $\gp$ is consistent, i.e.,
there exists a solution for $\gp$;

\item decide whether
a given set of hypotheses is an admissible solution for $\gp$;

\item decide whether a given set of hypotheses is
an optimal solution for $\gp$;

\item  decide whether a given hypothesis $h$ is relevant for $\gp$, i.e.,
$h$ occurs in some optimal solution of $\gp$;

\item decide whether a given hypothesis $h$ is necessary for $\gp$,
i.e., $h$ is contained in all optimal solutions of $\gp$;

\item compute an optimal solution of $\gp$.
\end{itemize}

The table in Figure \ref{fig:complexity} shows the complexity of all
these problems, both in the general case and in the restricted setting
where the use of unstratified negation is forbidden in the logic
program of the abduction problem.
Note that a complexity class $C$ in any entry of this table means that the
corresponding problem is $C$-complete, that is,
we prove both membership and hardness of the problem for the complexity class
$C$.

An interesting result in this course is that
``negation comes for free'' in most cases.
That is, the addition of negation does not cause any further
increase to the complexity of the main abductive reasoning tasks
(which remains the same as for not-free programs).
Thus, the user can enjoy the knowledge representation power
of nonmonotonic negation without paying additional costs in terms
of computational overhead.
More precisely, it turns out that abduction with
penalization over general logic programs has exactly the same
complexity as abduction with penalization over definite Horn
theories of classical logics in the three main computational
abductive-reasoning tasks
(deciding relevancy and necessity of an hypothesis,
and computing an optimal solution).
While unstratified negation brings a relevant complexity gap in deductive
reasoning (from $\gp$ to $\NP$ for brave reasoning),
in this case, the use of negation does not lead to any
increase in the complexity,
as shown in Figure \ref{fig:complexity}.

\begin{figure}[h]
\begin{tabular}{lcc}
  & General programs & Positive or stratified programs \\ \hline
  Consistency & $\NP$ & $\NP$ \\
  Solution Admissibility & $\NP$ & $\Pol$ \\
  Solution Optimality & $\DP{2}$ & $\CONP$ \\
  Hypothesis Relevancy & $\DeltaP{2}$ & $\DeltaP{2}$ \\
  Hypothesis Necessity & $\DeltaP{2}$ & $\DeltaP{2}$ \\
  Optimal Solution Computation & $\FPNP$ & $\FPNP$ \\ \hline
\end{tabular}
\caption{Overview of the Complexity Results}
\label{fig:complexity}
\end{figure}

We have implemented the proposed framework for abduction with penalization over logic programs as a front-end for the \dlv\ system.
Our implementation is based on an
algorithm that translates an abduction problem with penalties into
a logic program with weak constraints \cite{bucc-etal-99b}, which
is then evaluated by \dlv. We prove that our approach is sound and
complete. Our abductive system is available in the current release
of the \dlv\ system ({\tt www.dlvsystem.com}), and can be freely
retrieved for experiments.
It is worthwhile noting that our rewriting approach can be adapted
for other ASP systems with suitable constructs for
dealing with weighted preferences. For instance, our algorithm
can be modified easily in order to compute programs with weight literals
to be evaluated by the Smodels system \cite{simo-etal-2002}.

\vspace{0.2cm}

In sum, the main contribution of the paper is the following.
\begin{itemize}
\item
We define a formal model of abduction with penalization over logic programs.
\item
We carry out a thorough analysis of the complexity of the
main computational problems arising in the context of abduction
with penalization over logic programs.
\item
We address knowledge representation issues, showing how some relevant
problems can be encoded in our framework in a simple and fully declarative way.
\item
We provide an implementation of the proposed abductive framework on top of the \dlv\ system.
\end{itemize}

Our work is evidently related to previous studies on semantic and
knowledge representation aspects of abduction over logic programs.
In Section \ref{sec:related}, we discuss
the relationships of this paper with such previous studies and with
some further related issues.

The rest of the paper is organized as follows. In Section
\ref{sec:prelim}, we recall the syntax of (function-free) logic
programs and the stable model semantics. In Section \ref{sec:abd},
we define our model of abduction with penalization from logic
programs, and in Section \ref{sec:kr} we give some examples of
applications of this form of abduction in different domains. In
Section \ref{sec:complexity}, we analyze the computational
complexity of the main problems arising in this framework. In
Section \ref{sec:implementation}, we describe our prototype that
implements abduction with penalization from logic programs, and
makes it available as a front end of the system \dlv. Section
\ref{sec:related} is devoted to related works. Finally, in Section
\ref{sec:conclusion}, we draw our conclusions.

\section{Preliminaries on Logic Programming} \label{sec:prelim}

We next give the syntax of function-free logic programs, possibly
containing nonmonotonic negation (negation as failure) and
constraints.
Then, we recall the stable model semantics \cite{gelf-lifs-88} for
such logic programs.

\subsection{Syntax}

A $term$ is either a constant or a variable\footnote{ Note that
function symbols are not considered in this paper.}. An $atom$ has the form
$\mathtt{a(t_{1},..., t_{n})}$, where $\mathtt a$ is a $predicate$
of arity $n$ and $\mathtt{t_{1},..., t_{n}}$ are terms. A
$literal$ is either a $positive~literal$ $\mathtt a$ or a
$negative$ $literal$ $not \ \mathtt a$, where $\mathtt a$ is an
atom.

A {\em rule} $r$ has the form
\[
\mathtt{ a \derives b_1,\ldots, b_k, not \
        b_{k+1},\ldots, not \ b_m. \quad\quad k\geq 0, \ m\geq k
}\] where $\mathtt{ a ,b_1,\ldots ,b_m}$ are atoms.\\ Atom
$\mathtt a$ is the {\em head} of $r$, while the conjunction
$\mathtt{b_1 ,\ldots, b_k, not \ b_{k+1} ,\ldots,not \ b_m}$ is
the {\em body} of $r$. We denote by $H(r)$ the head atom $\mathtt
a$, and by $B(r)$ the set $\mathtt{ \{b_1 ,\ldots, b_k, not \
b_{k+1} ,\ldots,}$ $\mathtt{ not \ b_m \}}$ of the body literals.
Moreover, $B^+(r)$ and $B^-(r)$ denote the set of positive and
negative literals occurring in $B(r)$, respectively. If
$B(r)=\emptyset$, i.e., $m=0$, then $r$ is a {\em fact}.

A \textit{strong constraint} (integrity constraint) has the form
$\mathtt{\derives L_1,\ldots, L_m.}$, where each $\mathtt{L_i}$,
$\mathtt 1 \leq \ i \leq m$ , is a literal; thus, a strong
constraint is a rule with empty head.

A {\em (logic) program} $P$ is a finite set of rules and
constraints. A negation-free program is called {\em positive
program}. A positive program where no strong constraint occurs is
a {\em constraint-free} program.

A term, an atom, a literal, a
rule or a program is $ground$ if no variable appears in it. A ground
program is also called a {\em propositional} program.

\subsection{Stable model semantics}

Let $P$ be a program. The {\em Herbrand Universe} $U_{P}$ of $P$
is the set of all constants appearing in $P$. The {\em Herbrand
Base} $B _{P}$ of $P$ is the set of all possible ground atoms
constructible from the predicates appearing in the rules of $P$
and the constants occurring in $U _{P}$ (clearly, both $U_{P}$ and
$B _{P}$ are finite). Given a rule $r$ occurring in a program $P$,
a {\em ground instance} of $r$ is a rule obtained from $r$ by
replacing every variable $X$ in $r$ by $\sigma (X)$, where
$\sigma$ is a mapping from the variables occurring in $r$ to the
constants in $U_{P}$. We denote by $ground( P)$ the (finite) set
of all the ground instances of the rules occurring in $P$. An {\em
interpretation} for $P$ is a subset $I$ of $\BP$ (i.e., it is a
set of ground atoms). A positive literal $a$ (resp.\ a negative
literal $not \ a$) is true with respect to an interpretation $I$
if $a\in I$ (resp.\ $a \notin I$); otherwise it is false. A ground
rule $r$ is {\em satisfied} (or {\em true}) w.r.t. $I$ if its head
is true w.r.t. $I$ or its body is false w.r.t. $I$.

A {\em model} for $P$ is an interpretation $M$ for $P$ such that
every rule $r \in ground(\gp)$ is true w.r.t. $M$. If $P$ is a
positive program and has some model, then $P$ has a (unique) least
model (i.e., a model included in every model), denoted by
$lm(\gpd)$.

Given a logic program $P$ and an interpretation $I$, the {\em
Gelfond-Lifschitz transformation} of $P$ with respect to $I$ is
the logic program $P^I$ consisting of all rules $\mathtt{ a
\derives b_1,\ldots, b_k}$ such that $\ \ (1)\; \mathtt{a \derives
b_1,\ldots, b_k, not \ b_{k+1},\ldots, not \ b_m} \in P\quad
\mbox{and}\quad (2)\; \mathtt{b_i} \notin I,  \mbox{\ for all $k<i
\leq m$}.$\\
Notice that $not $ does not occur in $P^I$, i.e., it is a positive
program.

An interpretation $I$ is a {\em stable model} of $P$ if it is the
least model of its Gelfond-Lifschitz w.r.t. $I$, i.e., if $I =
lm(P^I)$ \cite{gelf-lifs-88}. The collection of all stable models
of $P$ is denoted by $\SM(P)$ (i.e., $\SM(P) = \set{I\ |\  I =
lm(P^I)}$).

\begin{example}\label{ex-stabmod}
Consider the following (ground) program $P$:
\[
\begin{array}{llll}
{\tt \lrule{a}{not \ b}} &\ \ \ {\tt \lrule{b}{not \ a}}&\ \ \ {\tt\lrule{c}{a}}&\ \ \
{\tt \lrule{c}{b}}
\end{array}
\]
The stable models of $P$ are $M_1 = \{\tt a,c\}$ and $M_2 = \{\tt
b,c\}$. Indeed, by definition of Gelfond-Lifschitz transformation,
$$P^{M_1} = \{{\tt \ a\derives, c\derives a,\ c\derives b\ \}} \ \
{\rm and}  \ \ P^{M_2} = \{{\tt \ b\derives, c\derives a,\
c\derives b\ \}}$$ and it can be immediately recognized that
$lm(P^{M_1})= M_1$ and $lm(P^{M_2}) = M_2$.
\end{example}

We say that an atom $p$ depends on an atom $q$ if there is a rule
$r$ in $P$ such that $p=H(r)$ and either $q\in B^+(r)$ or $ not \
q\in B^-(r)$. Let $\preceq$ denote the transitive closure of this
dependency relationship. The program $P$ is a {\em recursive}
program if there are $p,q\in\BP$ such that $p\preceq q$ and
$q\preceq p$. We say that $P$ is \emph{unstratified}, or that
unstratified negation occurs in $P$, if there is a rule $r$ in $P$
such that $p=H(r)$, $not \ q\in B^-(r)$, and $q\preceq p$. A
program where no unstratified negation occurs is called
\emph{stratified}.

Observe that every stratified program $P$ has at most one stable
model. The existence of a stable model is guaranteed if no strong
constraint occurs in the stratified program $P$. Moreover, every
stratified program can be evaluated in polynomial time. In
particular, deciding whether there is a stable model, computing
such a model, or deciding whether some literal is entailed (either
bravely or cautiously) by the program are all polynomial-time
feasible tasks.

For a set of atoms $X$, we denote by $\facts(X)$ the set of facts
$\{ p. \mid p\in X\}$. Clearly, for any program $P$ and set of
atoms $S$, all stable models of $P\cup\facts(S)$ include the atoms
in $S$.

\section{A Model of Abduction with Penalization} \label{sec:abd}

First, we give the formal definition of a problem of abduction from
logic programs under the stable model semantics, and
we provide an example on network diagnosis, that we use as a running example
throughout the paper.
Then, we extend this framework by introducing the notion of penalization.

\begin{definition}\label{def:abd-sol}
{\sc (Abduction From Logic Programs)}\\ A problem of abduction
from logic programs $\gp$ is a triple $\tuple{H,P,O}$, where $H$
is a finite set of ground atoms called {\em hypotheses}, $P$ is a
logic program whose rules do not contain any hypothesis in their
heads, and $O$ is a finite set of ground literals, called {\em
observations}, or {\em manifestations}.

A set of hypotheses $S\subseteq H$ is an {\em admissible solution}
(or {\em explanation}) to $\gp$ if there exists a stable model $M$
of $P \cup \facts(S)$ such that, $\forall o\in O$, $ o$ is true
w.r.t. $M$.

The set of all admissible solutions to $\gp$ is denoted by
$\ADM(\gp)$.  \qed
\end{definition}

\begin{example}\label{ex:network} {\sc (Network Diagnosis)}
{ Suppose that we are working on machine {\em a} (and we therefore
know that machine {\em a} is online) of the computer network
$\mathcal{N}$ in Figure~\ref{fig:network}, but we observe machine
{\em e} is not reachable from {\em a}, even if we are aware that
{\em e} is online. We would like to know which machines could be
offline. This can be easily modelled in our abduction framework
defining a problem of abduction $\gp_1=\tuple{H,P,O}$, where the
set of hypotheses is $H = {\tt \{offline(a),\ offline(b),}$ ${\tt
offline(c), offline(d),\ offline(e),\ offline(f)\}}$,\ the set of
observations is $O = {\tt \{ not \ offline(a),\ not \ offline(e),
not \ reaches(a,e)\}}$, and the program $P$ consists of the set of
facts encoding the network, $\facts( \{ {\tt connected(X,Y)}\mid
\{X,Y\}$ is an edge of $\mathcal{N}\})$, and of the following
rules:
\[
\begin{array}{lll}
\tt reaches(X,X) & \!\!\! \derives & \!\!\! \tt node(X), not \ \; offline(X). \\
\tt    reaches(X,Z) & \!\!\! \derives & \!\!\! \tt reaches(X,Y), \
connected(Y,Z),\ not \ \; {\tt offline}(Z).
\end{array}
\]

Note that the admissible solutions for $\gp_1$ corresponds to the
network configurations that may explain the observations in $O$.
In this example, $\ADM(\gp)$ contains five solutions
\begin{eqnarray*}
S_1 = &\{ \tt offline(f),\ offline(b)\}, \\
S_2 = &\{ \tt offline(f),\ offline(c), \ offline(d) \},\\
S_3 = &\{ \tt offline(f),\ offline(b), \ offline(c) \},\\
S_4 = &\{ \tt offline(f),\ offline(b), \ offline(d) \},\\
S_5 = &\{ \tt offline(f),\ offline(b),\ offline(c), \ offline(d) \}.
\end{eqnarray*}
\begin{figure}
\begin{center}
\epsfig{file=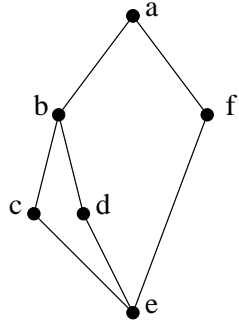,width=3cm}
\end{center}
\caption{Computer network $\mathcal{N}$ in Example \ref{ex:network}}
\label{fig:network}
\vspace{-0.5 cm}
\end{figure}
}
\end{example}

Note that Definition~\ref{def:abd-sol} concerns only the logical properties of
the hypotheses, and it does not take into account any kind of minimality
criterion. We next define the problem of abduction with penalization, which
allows us to make finer abductive reasonings, by expressing preferences on
different sets of hypotheses, in order to single
out the most plausible abductive explanations.

\begin{definition}{\sc (Abduction With Penalization From Logic Programs)}\label{def:opt-sol}\\
A {\em problem of abduction with penalization} (\PAP) $\gp$ is a
tuple $\tuple{H,P,O,\gamma}$, where $\tuple{H,P,O}$ is a problem
of abduction, and $\gamma$ is a polynomial-time computable
function from $H$ to the set of non-negative reals {\em (the
penalty function)}. The set of admissible solutions for $\gp$ is
the same as the set of solutions of the embedded abduction problem
$\tuple{H,P,O}$, i.e., we define $\ADM(\gp)= \ADM(\tuple{H,P,O})$.

For a set of atoms $A$,
let $sum_\gamma(A) = \sum_{h\in A} \gamma(h)$.
Then, $S$ is an {\em (optimal)
solution} (or {\em explanation}) for $\gp$ if (i) $S \in
\ADM(\gp)$ and (ii) $sum_\gamma(S) \leq sum_\gamma(S')$, for all
$S'\in \ADM(\gp)$.

The set of all (optimal) solutions for $\gp$ is denoted by
$\OPT(\gp)$.  \qed
\end{definition}

\begin{example}{\sc (Minimum-cardinality criterion)} \label{ex:cardinality}
Consider again the network $\mathcal{N}$ and the problem of
abduction $\gp_1=\tuple{H,P,O}$ in Example~\ref{ex:network}.
Again, we want to explain why the online machine $e$ is not
reachable from $a$. However, we do not consider any more plausible
all the explanations provided by $\gp_1$. Rather, our domain
knowledge suggests that it is unlikely that many machines are
offline at the same time, and thus we are interested in
explanations with the minimum number of offline machines. This
problem is easily represented by the problem of abduction with
penalization $\gp_2=\tuple{H,P,O,\gamma}$, where $H$, $P$ and $O$
are the same as in $\gp_1$, and, for each $h\in H$, $\gamma(h)=1$.

Indeed, consider the admissible solutions of $\gp_2$ and observe
that $$sum_\gamma(S_1)=2, \ \  sum_\gamma(S_2)= sum_\gamma(S_3)=
sum_\gamma(S_4)=3, \ \ sum_\gamma(S_5)=4$$ It follows that $S_1$
is the unique optimal explanation for $\gp_2$, and in fact
corresponds to the unique solution of our diagnosis problem with a
minimum number of offline machines.
\end{example}

The following properties of a hypothesis in a \PAP\ $\gp$ are of
natural interest with respect to computing abductive solutions.

\begin{definition}\label{defn:rel-nec}
 Let $\gpst$ be a \PAP\ and $h \in H$. Then, $h$ is {\em relevant}
for $\gp$ if $h \in S$ for some $S \in \OPT(\gp)$, and $h$ is
{\em necessary} for $\gp$ if $h \in S$ for every $S \in
\OPT(\gp)$.
\end{definition}

\begin{example}
In example \ref{ex:cardinality}, $\tt offline(b)$, and $\tt
offline(f)$ are the $relevant$ hypotheses; they are also
$necessary$ since $S_1$ is the only optimal solution.
\end{example}

\section{Knowledge Representation}\label{sec:kr}

In this section, we show how abduction with
penalization from logic programming can be used for encoding easily
and in a natural way relevant problems from different domains.

A nice discussion of how abduction can be used for
representing knowledge declaratively can be found in
\cite{dene-kaka-02}, where this setting is also related to other nonmonotonic
reasoning paradigms.
It also recalled that abduction has
been defined broadly as any form of ``inference to the best explanation"
\cite{jose-jose-94}, where {\em best} refers to the fact that usually
hypotheses can be compared according to some criterion.

In our framework, this optimality criterion is the sum
of the penalties associated to the hypotheses, which has to be minimized.

In particular, in order to represent a problem, we have to identify:
\begin{itemize}
\item {\em the hypotheses}, that represent all the possible entities
that are candidates for belonging to solutions;

\item for each hypothesis $h$, {\em the penalty} associated to $h$,
that represents the cost of including $h$ in a solution;

\item {\em the logic program} $P$, that encodes a representation of the
reality of interest and, in particular, of the way any given set of hypotheses
changes this reality and leads to some consequences;

\item {\em the observations}, or manifestations, that are distinguished
logical consequences, often encoding some {\em desiderata}.
For any given set of hypotheses $H$, the fact that these
observations are consequences of the logic program (plus $H$) witnesses
that $H$ is a ``good" set of hypotheses, i.e., it encodes a feasible solution for
the problem at hand.
\end{itemize}

For instance, in the network diagnosis problem described in Example \ref{ex:cardinality},
the hypotheses are the possible offline machines and the logic program $P$
is able to determine, for any given set of offline machines encoding a network
status, which machines are unreachable.
In this case, and usually in diagnosis problems, these observations are in fact
pictures of the reality of interest: we see that some machines are not reachable in the
network and that some machines are not offline, and we would like to infer,
via abductive reasoning, what are the explanations for such a situation.
Moreover, in this example, we are interested only in solutions that consist
of the minimum number of offline machines, leading to the observed
network status, because they are believed more likely to occur.
This is obtained easily, by assigning a unitary penalty to each
hypothesis.

We next show the encodings of other kind of problems that
can be represented in a natural way trough abduction with penalization from logic programs,
even though they are quite different from the above simple cause-effect scheme.

For the sake of presentation, we assume in this section that logic
programs are equipped with the built-in predicates $\neq$, $<$,
$>$, and $+$, with the usual meaning. Clearly, for any given
program $P$, these predicates may be encoded by suitable finite
sets of facts, because we have to deal only with the (finite) set
of constants actually occurring in $P$. Moreover, observe that
most available systems for evaluating logic programs -- e.g.,
\dlv\ \cite{eite-etal-98a,leon-etal-2002-dlv} and {\it smodels}
\cite{niem-simo-97,simo-etal-2002} -- provide in fact such
operators.

\subsection{The Travelling Salesman Problem} \label{sec:tsp}

An instance $I$ of the {\em Travelling Salesman Problem (TSP)}
consists of a number of cities $c_1,\ldots,c_n$, and a function $w$ that
assigns to any pair of cities $c_i,c_j$ a positive integer value,
which represents the cost of travelling from $c_i$ to $c_j$.
A solution to $I$ is a round trip that visits all cities in
sequence and has minimal travelling cost, i.e., a permutation
$\tau$ of $1,\ldots,n$ such that the overall cost
\[w(\tau) = \sum_{i=1}^{n-1} w(\tau(i),\tau(i+1)) + w(\tau(n),\tau(1))\]
is minimum.

Let us see how we can represent this problem in our framework.
Intuitively, any solution consists of pair of cities encoding a tour of
the salesman, while the observations must witness that this tour is correct,
i.e., that all cities are visited exactly once.
Thus, we have a hypothesis for each pair of cities $c_i$, $c_j$,
because any such a pair is candidate for belonging to
the trip of the salesman. The penalty associated to each hypothesis is
clearly the cost of travelling from $c_i$ to $c_j$, because we want to
obtain the minimum-cost tour.
Moreover, for any given trip encoded by a set of hypotheses,
the logic program determines the cities reached by the salesman, and
also whether the salesman has travelled in a correct way.
The observations are possible consequences of the program, which encode that
all cities are visited and no visiting rule has been violated.

Formally, we represent the TSP instance $I$ as a $\PAP$ $\gpst$
defined as follows. The set of hypotheses is $H= \{ \tt c(i,j)\ |\
1 \leq \tt i,\tt j \leq n\}$, where $\tt c(i,j)$ encodes the fact
that the salesman visits city $j$ immediately after city $i$. The
penalty function $\gamma( {\tt c(i,j)} ) = w({\tt i,j} )$ encodes
the cost of travelling from $i$ to $j$. The cities are encoded
through a set of atoms $\{ \tt city(i) \mid 1\leq \tt i\leq n\}$.
The program $P$ contains the following rules: {
\[
\begin{array}{lrlll}
(1) & \tt city(i). & & & \mbox{ for each } \tt i,\  1\leq \tt i \leq n \\
(2) & \tt visited(I) & \tt \derives & \tt {visited(J), c(J,I).}  &  \\
(3) & \tt visited(1) & \tt \derives & \tt c(J,1). &  \\
(4) & \tt missedCity  & \tt \derives & \tt {city(I), not \ visited(I).} & \\
(5) & \tt badTour & \tt \derives & \tt{ c(I,J), c(I,K), J \neq K.} &\\
(6) &  \tt  badTour &\tt \derives & \tt {c(J,I), c(K,I), J \neq K.} &\\
\end{array}
\]
} The observations are $O =\{\tt not \ \tt missedCity, not \ \tt
badTour \}$.

It is easy to see that every optimal solution $S\in \OPT(\gp)$
corresponds to an optimal tour and viceversa. The facts (1) of $P$
encode the cities to be visited. Rule (2) states that a city $i$
has been visited if the salesman goes to city $i$ after an already
visited city $j$. Rule (3) concerns the first city that, w.l.o.g.,
is the first and the last city of the tour. In particular, it is
considered visited, if it is reached by some other city $j$, which
is turn forced to be visited, by the other rule of $P$. Rule (4)
says that there is a missed city if at least one of the cities has
not been visited. Atom $\tt badTour$, defined by rules (4) and
(5), is true if some city is in two or more connection endpoints
or connection startpoints. The observations $\tt not \
missedCity,not \ badTour$ enforce that admissible solutions
correspond to salesman tours that are complete (no city is missed)
and legal (no city is visited twice).

Moreover, since optimal solutions
minimize the sum of the connection costs, abductive solutions in
$\OPT(\gp\ )$ correspond one-to-one to the optimal tours.


In \cite{eite-etal-97f}, Eiter, Gottlob, and Mannila show that
Disjunctive Logic Programming (function-free logic programming
with disjunction in the heads and negation in the bodies of the
rules) is highly expressive.
Moreover, the authors strength the theoretical analysis of the
expressiveness by proving that problems relevant in practice like,
e.g., the {\em Travelling Salesman Problem} and {\em
Eigenvector}, can be programmed in DLP, while they cannot be
expressed by disjunction-free programs.
Indeed, recall that computing an optimal tour is both \NP-hard and \CONP-hard.
Moreover, in \cite{papa-84} it is shown that deciding whether the cost of an
optimal tour is even, as well as deciding whether there exists a unique
optimal tour, are $\DeltaP{2}$-complete problems.
 Hence, it is not possible to express this problem in
 disjunction-free logic programming, even if unstratified
 negation is allowed (unless the polynomial hierarchy collapses).

Nevertheless, the logic
programs implementing these problems in DLP highlight, in our
opinion, a weakness of the language for the representation of
optimization problems. The programs are very complex and tricky,
the language does not provide a clean and declarative
way to implement these problems.%
\footnote{We refer to standard Disjunctive Logic Programming here.
As shown in \cite{bucc-etal-99b}, the addition of {\em weak
constraints}, implemented in the \dlv\ system
\cite{eite-etal-2000c}, is another way to enhance DLP to naturally
express optimization problems.}
For a comparison, we report in Appendix \ref{app:tsp}
the encoding of this problem in (plain) DLP, as
described in \cite{eite-etal-97f}.
Evidently, abduction with penalization provides a
simpler, more compact, and more elegant encoding of TSP.
Moreover, note that, using this form of abduction, even normal
(disjunction-free) programs are sufficient for encoding such
optimization problems. \nop{ and make logic programming a valuable
and useful tool for expressing wide class of problems. }

\nop{ Observe that the logic program $LP$ above consists of only
4 rules (that do not even contain disjunction and negation);
while the program of Example \ref{exa-tsp} consists of 20 rules. }

\subsection{Strategic Companies}

We present a new version of the {\em strategic companies} problem
\cite{cado-etal-97}.
A manager of a holding identifies a set of
crucial goods, and she wants these goods to be produced by the
companies controlled by her holding. In order to meet this goal,
she can decide to buy some companies, that is to buy enough shares
to get the full control of these companies.
Note that, in this
scenario, each company may own some quantity of shares of another
company. Thus, any company may be controlled  either directly, if
it is bought by the holding, or indirectly, through the control
over companies that own more than $50\%$ of its shares. Of course,
it is prescribed to minimize the quantity of money spent for
achieving the goal, i.e., for buying new companies.

For the sake of simplicity, we will assume that, if a company $X$ can
be controlled indirectly, than there are either one or two companies that
together own more than $50\%$ of the shares of $X$.
Thus, controlling these companies is sufficient to take the control
over $X$.

We next describe a problem of abduction from logic programs with
penalization $\gp$ whose optimal solutions correspond to the optimal choices
for the manager.
In this case, the observations are the crucial goods that we want to produce,
while the hypotheses are the acquisitions of the holding and their associated
penalties are the costs of making these financial operations.
The logic program determines, for any given set of
acquisitions, all the companies controlled by the holding and
all the goods produced by these companies.

Companies configurations are encoded by the set of atoms {\it
Market} defined as follows: if a company $y$ owns $n\%$ of the
shares of a company $x$ then ${\tt share(x,y,n)}$ belongs to {\it
Market}, and if a company $x$ produces a good $a$ then ${\tt
producedBy(a,x)}$ belongs to {\it Market}. No more atoms belong to
this set.

Then, let $\gpst$, where the set of hypotheses $H$ = \{$\tt
bought(x_1),\dots,$ $\tt bought(x_n)$\} encodes the companies that
can be bought, and the set of observations $O =\{\tt
produced(y_1),\dots,produced(y_n) \}$ encodes the set of goods to
be produced. Moreover, for each atom ${\tt bought(\bar x)} \in H$,
$\gamma({\tt bought(\bar x)})$ is the cost of buying the company
$\bar x$. The program $P$ consists of the facts encoding the state
of the market $\facts({\it Market})$ and of the following rules:

\newcounter{fup}
\def\nbr{\pr{fup}}
\[
\begin{array}{l@{\hspace*{0.5cm}}rcll}
\nbr & \tt produced(X) & \tt \derives & \tt{producedBy(X,Y), controlled(Y).} \\
\nbr & \tt controlled(X) &\tt \derives & \tt bought(X). \\
\nbr & \tt controlled(X) &\tt \derives & \tt {share(X,Y,N), controlled(Y), N > 50.}\\
\nbr & \tt controlled(X) &\tt \derives & \tt {share(X,Y,N), share(X,Z,M),} \\
     &               &         & \tt {  controlled(Y), controlled(Z), M + N > 50, Y\neq Z.}\\
\end{array}
\]
\vspace{1cm}
\begin{figure}[h]
\begin{center}
\epsfig{file=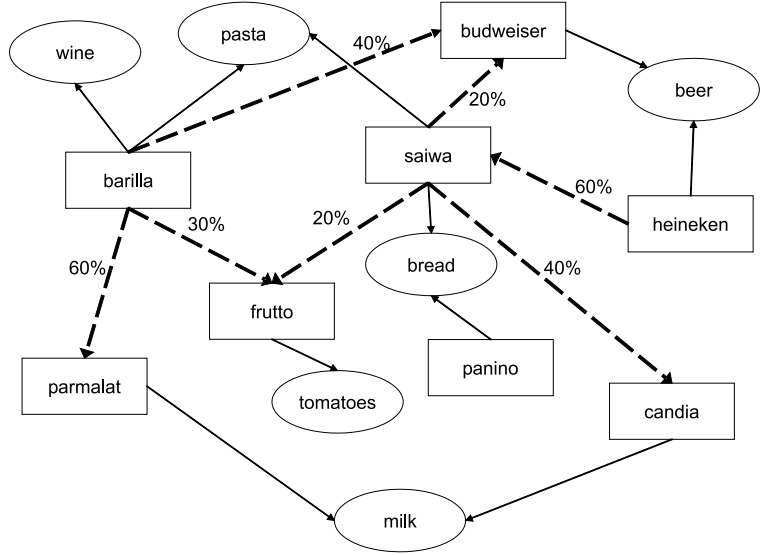,width=10cm}
\end{center}
\caption{Strategic Companies} \label{fig:strategic} \vspace{-0.5
cm}
\end{figure}

\begin{example}
Consider the following sets of companies and goods:\\
{\it Companies}=\{ {\it barilla, saiwa, frutto, panino, budweiser,
heineken,
parmalat, candia} \} \\
{\it Goods}= \{ {\it wine, pasta, beer, tomatoes,
bread, milk}\}. \\
Figure \ref{fig:strategic} depicts the relationships $\tt share$
and $\tt producedBy$ among companies, and among products and
companies, respectively. A solid arrow from a company $C$ to a
good $G$ represents that $G$ is produced by $C$. A dashed arrow
from a company $C_1$ to a company $C_2$ labelled by $n$ means that
$C_1$ owns $n\%$ of the shares of $C_2$. The cost (in millions of
dollars) for buying directly a company is shown below:
\[
\begin{array}{ll@{\hspace{0.7cm}}ll@{\hspace{0.7cm}}ll@{\hspace{0.7cm}}ll}
 barilla & 500 & saiwa & 400 & frutto & 350 & panino & 150\\
 budweiser & 300 & heineken & 300 & parmalat & 300 & candia & 150\\
\end{array}
\]

Accordingly, the hypotheses and their respective penalties are
\[
\begin{array}{lll}
& \tt bought(barilla) & \gamma(\tt{bought(barilla)})=500\\
& \tt bought(saiwa) & \gamma( \tt{bought(saiwa)})= 400\\
&  & \cdots \\
& \tt bought(candia) & \gamma(\tt{ bought(candia)})=150.\\
\end{array}
\]

The set of observations is
\[
\begin{array}{ll@{\hspace{-0.3pt}}l}
& O = & \{{\tt produced(pasta)}, {\tt
produced(wine)}, {\tt produced(tomatoes)},\\
&  & \ \ {{\tt produced(bread)},\tt produced(beer)}, {\tt produced(milk)}\}\\
\end{array}
\]

This problem has the only optimal solution $S_1$=\{{\it barilla,
frutto, heineken}\}, whose cost is 1150 millions of dollars. Note
that all goods in $G$ can be produced also by buying the set of
companies $S_2$= \{{\it barilla, frutto, saiwa}\}. However, since
$saiwa$ is more expensive than $heineken$, $S_1$ is preferred to
$S_2$.
\end{example}

\subsection{Blocks world with penalization}

Planning is another scenario where abduction proves to be useful
in encoding hard problems in an easy way.

The topic of logic-based languages for planning
has recently received a renewed great deal of interest,
and many approaches based on answer set semantics, situation calculus, event calculus,
and causal knowledge have been proposed
--- see, e.g., \cite{gelf-lifs-98,eite-etal-2001d,shan-2000,turn-99}.

We consider here the Blocks World Problem {\cite{lifs-99a}:
given a set of blocks $B$ in some initial configuration ${\it Start}$,
a desired final configuration ${\it Goal}$,
and a maximum amount of time {\it lastTime},
find a sequence of moves leading the blocks from state ${\it Start}$,
to state ${\it Goal}$ within the prescribed time bound.
Legal moves and configurations obey the following rules:
A block can be either on the top of another block, or on the table.
A block has at most one block over it, and is said to be clear
if there is no block over it.
At each step (time unit), one or more clear blocks can be moved either on
the table, or on the top of other blocks.
Note that in this version of the Blocks World Problem more than one
move can be performed in parallel, at each step.
Thus, we additionally require that a block $B_1$ cannot be moved
on a block $B_2$ at time $T$ if also $B_2$ is moved at time $T$.

Assume that we want to compute legal sequences of moves -- also called plans
-- that leads to the desired final state within the {\it lastTime} bound
and that consists of the minimum possible number of moves.
We next describe a problem of abduction from logic programs with
penalization $\gp$ whose optimal solutions correspond to such good plans.
In this case, the observations encode the desired final configuration,
while the hypotheses correspond to all possible moves.
Since we are interested in minimum-length plans, we assign a unitary penalty
to each move (hypothesis).
Finally, the logic program has to determine the state of the system
after each move and detect possible illegal moves.

Consider an instance BWP of the Blocks World Problem.
Let $B=\{b_1,\cdots,b_n\}$ be the set of blocks and
$L=B\cup\{{\it table}\}$ the set of possible locations.

The blocks of BWP are encoded by the set of atoms ${\it Blocks}=$
$\{{\tt block(b_1),\ldots, block(b_n)}\}$, and the initial
configuration is encoded by a set {\it Start} containing atoms of
the form $\tt on(b,\ell,0)$, meaning that, at time $0$, the block
$b$ is on the location $\ell$.

Then, let $\gpst$. The set of hypotheses $H = \{ {\tt
move(b,\ell,t)} \mid b \in B,\ell \in L, 0\leq t < {\it
  lastTime}\}$ encodes all the
possible moves, where an atom $\tt move(\bar b,\bar\ell,\bar t)$
means that, at time $\bar t$, the block $\bar b$ is moved to the
location $\bar\ell$. The set of observations $O$ contains atoms of
the form $\tt on(b,\ell,{\it lastTime})$, encoding the final
desired state {\it Goal}. The penalty function $\gamma$ assigns
$1$ to each atom ${\tt move(b,\ell,t)} \in H$. Moreover, $P=
\facts({\it Blocks})\cup \facts({\it Start})\cup R$, where $R$ is
the following set of rules:
\begin{eqnarray}
       \tt on(B,L,T1) & \tt \derives & \tt {move(B,L,T), T1=T+1.} \label{rule3}\\
[-3pt] \tt on(B,L,T1) & \tt \derives & \tt {on(B,L,T), T1=T+1, not \ moved(B,T).} \label{rule4}\\
[-3pt] \tt moved(B,T) & \tt \derives & \tt {move(B,\_,T).}\label{rule5}\\
[-3pt] & \tt \derives & \tt {on(B,L,T), on(B,L1,T), L \not= L1.} \label{rule6}\\
[-3pt] & \tt \derives & \tt {on(B1,B,T),on(B2,B,T),B2\not= B1,block(B).} \label{rule13}\\
[-3pt] & \tt \derives & \tt {on(B,B,T).} \label{rule7}\\
[-3pt] & \tt \derives & \tt {move(B,B1,T), move(B1,L,T).} \label{rule9}\\
[-3pt] & \tt \derives & \tt {move(B,L,T), on(B1,B,T), B \not= B1.}
\label{rule11}
\end{eqnarray}

Note that the strong constraints in $R$ discards models encoding
invalid states and illegal moves.
For instance, Constraint \ref{rule11} says that it is forbidden
to move a block $B$, if $B$ is not clear.

Rule \ref{rule3} says that moving a block
$B$ on a location $L$ at time $T$ causes $B$ to be on $L$ at time $T+1$.
Rule \ref{rule4} represents the {\em inertia} of blocks, as it asserts
that all blocks that are not moved at some time $T$
remain in the same position at time $T+1$.

It is worthwhile noting that expressing such {\em inertia rules}
is an important issue in knowledge representation, and clearly
shows the advantage of using logic programming, when nonmotononic
negation is needed.

For instance, observe that Rule \ref{rule4} is very
natural and intuitive, thanks to the use of negation in literal
${\tt not \ moved(B,T)}$. However, it is not clear how to express
this simple rule -- and inertia rules in general -- by using
classical theories.%
\footnote{In fact, there are some solutions to this problem for
interesting special cases, such as settings where all actions on
all fluents can be specified \cite{reit-91}.
Also, in \cite{mcca-turn-97}, it is defined a nonmonotonic formalism
based on causal laws that
is powerful enough to represent inertia rules (unlike previous approaches
based on inference rules only).
A comprehensive discussion of the frame problem can be found in
the book \cite{shan-97}.
}

\begin{figure}[h]
\begin{center}
\epsfig{file=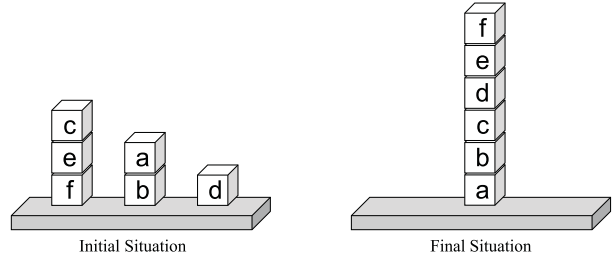,width=8 cm}
\end{center}
\caption{Blocks World} \label{fig:bw}
\end{figure}

\begin{example}
Consider a Blocks World instance where the initial configuration
and the final desired state are shown in figure \ref{fig:bw}, and
the maximum number of allowed steps is $6$. Therefore, the set of
observations of our abduction problem is $\{ {\tt
on(a,\,table,\,6)}$, $\tt on(b,\,a,\,6)$, $\tt on(c,\,b,\,6)$,
$\tt on(d,\,c,\,6)$, $\tt on(e,\,d,\,6)$, $\tt on(f,\,e,\,6)\}$.
The set of hypotheses contains all the possible moves, that is
\vspace{-0.6 cm}

$$H
= \{\tt move(a,\,table,\,0), \ \tt move(a,\,table,\,1), \ \cdots,
\ \tt move(f,\,d,\,6), \ \tt move(f,\,e,\,6)\}$$ \vspace{-0.6 cm}

 Each move has
cost $1$.

In this case, the minimum number of moves needed for reaching the
final configuration is six. An optimal solution is \{$\tt
move(a,\,table,\,0)$, $\tt move(b, \,a,\,1)$, $\tt
move(c,\,b,\,2)$, $\tt move(d,\,c,\,3)$, $\tt move(e,\,d,\,4)$,
$\tt move(f, \,e, \,5)$\}. Note that the plan
\[
\begin{array}{ll}
& \{\tt move(a,\,table,\,0), \tt move(c,\,table,\,0), \tt
move(b,\,a,\,1), \\
& \ \ \tt move(c, \,b,\,2), \tt move(d,\,c,\,3), \tt
move(e,\,d,\,4), \tt
move(f,\,e,\,5)\}  \\
\end{array}
\]

though legal, is discarded by the minimality criterion, because it
consists of seven moves.
\end{example}

Finally, observe that the proposed framework of abduction from
logic programs with penalties allows us to represent easily
different plan-optimization strategies. For instance, assume that
each block has a weight, and we want to minimize the total effort
made for reaching the goal. Then, it is sufficient to modify the
penalty function in the $\PAP$ $\gp$ above as follows: for each
hypothesis $\tt move(b,\ell,t)$, let $\gamma({\tt
move(b,\ell,t)})=w$, where $w$ is the weight of the block $b$.

\section{Computational Complexity}\label{sec:complexity}

In this section, we study the computational complexity of the main
problems arising in the framework of abduction with
penalization from logic programs, both in the general case
and when some syntactical restrictions are placed on logic programs.

\subsection{Preliminaries on Complexity Theory}
\label{sec:prelcomplexity}

For \NP-completeness and complexity theory, the reader is referred to
\cite{papa-94}. The classes $\SigmaP{k}, \PiP{k}$ and $\DeltaP{k}$ of the
Polynomial Hierarchy (PH) (cf.\ \cite{stoc-87}) are defined as follows:
\[
\begin{array}{cc}
 \DeltaP{0} = \SigmaP{0} = \PiP{0} = \Pol\quad \mbox{and for all $k \geq 1$,} \\
\DeltaP{k} = \Pol^{\Sigma_{k-1}^P},\;\; \SigmaP{k} =
\NP^{\Sigma_{k-1}^P},\;\; \PiP{k} = \mbox{co-}\SigmaP{k}.
\end{array}
\]
\noindent In particular, $\NP = \SigmaP{1}$, $\CONP = \PiP{1}$,
and $\DeltaP{2} = \Pol^{\NP}$. Here $\Pol^C$ and $\NP^C$ denote
the classes of problems that are solvable in polynomial time on a
deterministic (resp.\ nondeterministic) Turing machine with an
oracle for any problem $\pi$ in the class $C$. The oracle replies
to a query in unit time, and thus, roughly speaking, models a call
to a subroutine for $\pi$ that is evaluated in unit time. The
class $\DP{k}$ contains all problems that consist of the
conjunction of two (independent) problems from $\SigmaP{k}$ and
$\PiP{k}$, respectively. In particular, $\DP{2}$ is the class of
problems that are the conjunction of an $\NP$ and a $\CONP$
problem.

Notice that for all $k \geq 1$,
\[ \SigmaP{k} \subseteq\; \DP{k+1} \;\subseteq\; \DeltaP{k+1} \;\subseteq\; \SigmaP{k+1} \;\subseteq \; {\rm PSPACE}, \]
where each inclusion is widely conjectured to be strict.

We are also interested in the complexity of computing solutions, and
thus in classes of functions.
In particular, we consider the class $\FPNP$, which is the class of
functions corresponding to $\Pol^{\NP}$ ($\DeltaP{2}$), and
characterizing the complexity of many relevant optimization problems,
such as the TSP problem \cite{papa-84,papa-94}.
Formally, this is the class of all functions that can be computed by a
polynomial-time deterministic Turing transducer with an oracle in $\NP$.
Note that the only difference with the corresponding class of decision
problems is that deterministic Turing transducers are equipped with
an output tape, for writing the result of the computation.

\subsection{Complexity Results}
\label{sec:results}

Throughout this section, we consider problems $\gpst$ such that
$P$ is a ground program, unless stated otherwise.

Let $\Phi = \{C_1, \ldots, C_n\}$ be a CNF propositional formula
over variables $X_1,\ldots,X_r$, denoted by $\var(\Phi)$. With
each $X_i\in \var(\Phi)$, $1\leq i\leq r$, we associate two atoms
$x_i,\bar x_i$ (denoted by lowercase characters), and an auxiliary
atom ${\it assigned}_i$, representing the propositional variable
$X_i$, its negation $not \ X_i$, and the fact that some truth
value has been assigned to it, respectively. Moreover, with each
clause $C: \ell_1\vee \cdots \vee\ell_m$ in $\Phi$, we associate a
rule $r(C): {\it contr}\derives {\it negate}(\ell_1),\ldots ,{\it
negate}(\ell_m)$, where ${\it negate}(\ell) = \bar x$, if $\ell =
X$, and   ${\it negate}(\ell) = x$, if $\ell = not \ X$.

Define $P(\Phi)$ as the constraint-free positive program
containing the following rules:
\[
\begin{array}{ll}
 r(C_i). & 1\leq i\leq n\\
 {\it inconsistent} \derives x_j,\bar x_j. & 1\leq j\leq r\\
 {\it assigned}_j \derives x_j. & 1\leq j\leq r\\
 {\it assigned}_j \derives \bar x_j. & 1\leq j\leq r\\
 {\it allAssigned}\derives {\it assigned}_1,\ldots,{\it assigned}_r.
\end{array}
\]

Let $R$ be any set of rules whose heads are from
$\bigcup_{i= 1}^r \{x_i,\bar x_i\}$.
Note that, for any stable model $M$ of $P(\Phi)\cup R$,
${\it allAssigned}\in M$ and ${\it inconsistent}\notin M$ hold if and
only if, for each $X\in\var(\Phi)$, exactly one atom from $\{x,
\bar x\}$ belongs to $M$. That is, $M$ encodes a truth-value
assignment for $\Phi$. Moreover, ${\it contr}\notin M$ only if
such a truth-value assignment satisfies all clauses of the formula
$\Phi$. In this case, we say that $\Phi$ is satisfied by $M$.

On the other hand, given any truth-value assignment $T: \var(\Phi)
\rightarrow \{\true,\false\}$, we denote by $\at(T)$ the set of
atoms $\{ x \mid X\in\var(\Phi) \mbox{ and } T(X)=\true\} \cup \{
\bar x \mid X\in\var(\Phi) \mbox{ and } T(X)=\false\}$. It can be
verified easily that, if $T$ satisfies $\Phi$, then
$P(\Phi)\cup\facts(\at(T))$ has a unique stable model that
contains {\it allAssigned} and contains neither \contr\ nor
\incons.

The first problem we analyze is the consistency problem. That is the
problem of deciding whether a $\PAP$ has some solution.

\begin{theorem}\label{theo:consistency}
Deciding whether a $\PAP$ $\gpst$ is consistent is $\NP$-complete.
Hardness holds even if $P$ is a constraint-free positive program.
\end{theorem}
\begin{proof}
\noindent (\emph{Membership}).  We guess a set of hypotheses
$S\subseteq H$ and a set of ground atoms $M$, and then check that
(i) $M$ is a stable model of $P\cup \facts(S)$, and (ii) $O$ is
true w.r.t. $M$. Both these tasks are clearly feasible in
polynomial time, and thus the problem is in $\NP$.

\

\noindent (\emph{Hardness}). We reduce SAT to the consistency
problem. Let $\Phi$ be a CNF formula and $P(\Phi)$ its
corresponding logic program, as described above. Consider the
$\PAP$ problem $\langle H,P(\Phi), O,\gamma \rangle$, where $H =
\{x,\bar x \mid X\in \var(\Phi)\}$, $O = \{not \ {\it contr},$
$not \ {\it inconsistent},$ ${\it
  allAssigned}\}$,
and $\gamma$ is the constant function 0.

Let $S$ be an admissible solution for $P$, that is, there is a
stable model $M$ for $P(\Phi)$ such that {\it allAssigned} belongs
to $M$, and neither \contr\ nor \incons\ belongs to $M$. As
observed above, this entails that $\Phi$ is satisfied by the
truth-assignment corresponding to $M$, and in fact encoded by the
set of hypotheses $S$. Moreover, if $\Phi$ is satisfiable, there
is a truth-assignment $T$ that satisfies it. Then, it is easy to
check that $\at(T)$ is an admissible solution for $P$, since the
unique stable model of $P(\Phi)\cup\facts(\at(T))$ contains {\it
allAssigned} and no atom in $\{\contr,\incons\}$. Thus, $\Phi$ is
satisfiable if and only if $P$ is consistent. Finally, note that
$P$ can be computed in polynomial time from $\Phi$, and that $P$
does not contain negation or strong constraints.
\end{proof}

We next focus on the problem of checking whether a given set of
atoms $S$ is an admissible solution for a $\PAP$ $\gpst$. Observe
that this task is clearly feasible in polynomial time if $P$ is
{\it stratified}, because in this case the (unique) stable model
of $P\cup\facts(S)$ (if any, remember that strong constraints may
occur in $P$) can be computed in polynomial time. It follows that
this problem is easier than the consistency problem in this
restricted setting. However, we next show that it remains
$\NP$-complete, in the general case.

\begin{theorem}\label{theo:admissible}
Deciding whether a set of atoms is an admissible solution for a
$\PAP$ is $\NP$-complete.
\end{theorem}
\begin{proof}
\noindent (\emph{Membership}). Let $\gpst$ be a $PAP$ and $S$ a
set of atoms. We guess a set of ground atoms $M$, and then check
that (i) $M$ is a stable model of $P\cup \facts(S)$, and (ii) $O$
is true w.r.t. $M$. Both these tasks are clearly feasible in
polynomial time, and thus the problem is in $\NP$.

\

\noindent (\emph{Hardness}). We reduce SAT to the admissible
solution problem. Let $\Phi$ be a CNF formula over variables
$\{X_1,\ldots,X_r\}$, and $P(\Phi)$ its corresponding logic
program. Consider the $\PAP$ problem $P=\langle
\emptyset,P(\Phi)\cup G(\Phi), O,\gamma \rangle$, where $O = \{
not \ {\it contr}, not \ {\it inconsistent}\}$, $\gamma$ is the
constant function 0, and $G(\Phi)$ contains two rules $x\derives
not \ \bar x$ and $\bar x\derives not \ x$, for each
$X\in\var(\Phi)$.

Let $M$ be a stable model of $P(\Phi)\cup G(\Phi)$. Because of the
rules in $G(\Phi)$, for each pair of atoms $x,\bar x$ occurring in
it, either $x$ or $\bar x$ belongs $M$, and hence {\it
allAssigned}, too. Thus, these atoms encode a truth-assignment $T$
for $\Phi$. Moreover, it is easy to check that $\contr,\incons
\notin M$ only if this assignment $T$ satisfies $\Phi$. On the
other hand, let $T'$ be a satisfying truth-assignment for $\Phi$,
and let $M'= \at(T') \cup {\it allAssigned} \cup \{ {\it
assigned}_j \mid 1\leq j\leq r\}$. Then, $M'$ is a stable model of
$P(\Phi)$, and $\contr,\incons \notin M'$, that is, all
observations are true w.r.t. $M'$.

Therefore, $\emptyset$ is an admissible solution for $P$
if and only if $\Phi$ is satisfiable.
Note that unstratified negation occurs in $G(\Phi)$.
\end{proof}

It turns out that deciding whether a solution is optimal is both
$\NP$-hard and $\CONP$-hard.
However, this problem is not much more difficult than problems in these
classes, as we need to solve just an
$\NP$ and a $\CONP$-problem, independent of each other.

\begin{theorem}\label{theo:optimality} 
Deciding whether a set of atoms is an
optimal solution for a $\PAP$ is $\DP{2}$-complete.
\end{theorem}
\begin{proof}
(\emph{Membership}). Let $\gpst$ be a $\PAP$ and let $S$ be a set
of atoms. To prove that $S$ is an optimal solution for $\gp$ first
check that $S$ is an admissible solution, and then check there is
no better admissible solution. The former task is feasible in
$\NP$, by Theorem \ref{theo:admissible}. The latter is feasible in
$\CONP$. Indeed, to prove that there is an admissible solution
better than $S$, we guess a set of atoms $S'\subseteq H$ and a
model $M$ for $P$, and then check in polynomial time that
$sum_\gamma(S')< sum_\gamma(S)$, $M$ is a stable model of $P\cup
\facts(S')$, and $O$ is true w.r.t. $M$.

\

\noindent (\emph{Hardness}). Let $\Phi_1$ and $\Phi_2$ be two CNF
formulas, over disjoint sets of variables $\{X_1,\ldots,X_r\}$ and
$\{X'_1,\ldots,X'_v\}$. Deciding whether $\Phi_1$ is satisfiable
and $\Phi_2$ is not satisfiable is a $\DP{2}$-complete problem
\cite{papa-yann-84}. Let $P(\Phi_1)$ be the logic program
associated with $\Phi_1$, and $G_s(\Phi_1)$ a set of rules that
contains, for each $x\in\var(\Phi_1)$, two rules $x\derives not \
\bar x, s$ and $\bar x\derives not \ x, s$. Let $P'(\Phi_2)$ be
the logic program associated with $\Phi_2$, but for the atoms {\it
contr}, {\it inconsistent}, and {\it allAssigned}, which are
uniformly replaced in this program by ${\it contr}'$, ${\it
inconsistent}'$, and ${\it allAssigned}'$, respectively. Moreover,
let $R$ be the set containing two rules ${\it ok}\derives not \
{\it contr},not \ {\it inconsistent}, {\it allAssigned}$ and ${\it
ok}\derives not \ {\it contr}', not \ {\it inconsistent}',
   {\it allAssigned}'$.
Then, define $P(\Phi_1,\Phi_2)$ as the $\PAP$ problem $\langle
H,P,O,\gamma \rangle$, where $P= P(\Phi_1)\cup G_s(\Phi_1)\cup
P'(\Phi_2)\cup R$, $H= \{s\}\cup \{x',\bar x' \mid X'\in
\var(\Phi_2)\}$, $O = \{{\it ok}\}$, and the penalty function
$\gamma$ is defined as follows: $\gamma(s)=1$ and $\gamma(h)=0$,
for any other hypothesis $h\in H-\{s\}$.

We claim that $\Phi_1$ is satisfiable and $\Phi_2$ is not satisfiable
if and only if $\{s\}$ is an optimal solution for $P(\Phi_1,\Phi_2)$.

\emph{(Only if)}. Assume that $\Phi_1$ is satisfiable and $\Phi_2$
is not satisfiable, and let $T_1$ be a satisfying truth-value
assignment for $\Phi_1$. Moreover, let $M= \{\at(T_1) \cup \{ {\it
assigned}_j \mid 1\leq j\leq r\} \cup \{s, {\it allAssigned},
ok\}$. Then, $M$ is a stable model of $P \cup \facts(\{s\})$ and
thus $\{s\}$ is an admissible solution for $P(\Phi_1,\Phi_2)$, and
its cost is $1$, as $\gamma(s)=1$, by definition. Note that the
only way to reduce the cost to $0$ is by finding a set of
hypotheses that do not contain $s$, and is able to derive the
observation $ok$. From the rules in $R$, this means that we have
to find a subset of $\{x',\bar x' \mid X'\in \var(\Phi_2)\}$,
which encodes a satisfying truth assignment for $\Phi_2$. However,
this is impossible, because $\Phi_2$ is not satisfiable, and thus
$\{s\}$ is optimal.

\emph{(If)}.
Assume that $\{s\}$ is an optimal solution for $P(\Phi_1,\Phi_2)$.
Its cost is $1$, because $\gamma(s)=1$.
Note that any set of hypotheses $S'$ that encodes a satisfying
truth-value assignment for
$\Phi_2$ and does not contain $s$ is an admissible solution for
$P(\Phi_1,\Phi_2)$, and has cost $0$.
It follows that $\Phi_2$ is not satisfiable, as we assumed
$\{s\}$ is an optimal solution.
Therefore, by definition of $R$, the only way to derive the atom $ok$
is through the rule ${\it ok}\derives not \ {\it contr},not \ {\it
  inconsistent}, {\it allAssigned}$.
Since $\{s\}$ is also an admissible solution, we conclude that
there is a stable model $M$ that contains {\it allAssigned}, and no
atom from $\{\incons,$ $\contr\}$. That is, $M$ encodes a satisfying
truth assignment for $\Phi_1$.
\end{proof}

If unstratified negation does not occur in logic programs,
we lose a source of complexity, as
checking whether a solution is admissible is easy.
In fact, we show below that, in this case,
the optimality problem becomes $\CONP$-complete.

\begin{theorem}
Let $\gpst$ be a $\PAP$, where $P$ is a stratified program.
Deciding whether a set of atoms $S$ is an optimal solution for
$\gp$ is $\CONP$-complete. Hardness holds even if $P$ is a
constraint-free positive program.
\end{theorem}
\begin{proof}
(\emph{Membership}). Recall that checking whether a solution $S$
is admissible is feasible in polynomial time if $P$ is stratified.
Thus, we have to check only that there is no admissible solution
better than $S$, and this task is in $\CONP$, as shown in the
proof of Theorem \ref{theo:optimality}.

\

\noindent (\emph{Hardness}). Let $\Phi$ be a CNF formula,
$P(\Phi)$ its corresponding logic program, and $R$ be the set
containing two rules ${\it ok}\derives s$ and ${\it ok}\derives
{\it allAssigned}$. Then, define $P(\Phi)$ as the $\PAP$ problem
$\langle H,P,O,\gamma \rangle$, where $P= P(\Phi)\cup R$, $H=
\{s\}\cup \{x,\bar x \mid X\in \var(\Phi)\}$, $O = \{{\it ok}, not
\ {\it contr}, not \ {\it inconsistent}\}$, and the penalty
function $\gamma$ is defined as follows: $\gamma(s)=1$ and
$\gamma(h)=0$, for any other hypothesis $h\in H-\{s\}$.

We claim that $\Phi$ is not satisfiable
if and only if $\{s\}$ is an optimal solution for $P(\Phi)$.

\emph{(Only if)}. Assume $\phi$ is not satisfiable. Then, there is
no way of choosing a set of hypotheses that contains neither
\contr\ nor \incons\ and, furthermore, contains {\it allAssigned}
and hence  $ok$, but not $s$. It follows that the minimum cost for
admissible solutions is $1$. Moreover, note that $\{s\}$ is an
admissible solution for $P(\Phi)$, its cost is $1$, and thus it is
also optimal.

\emph{(If)}. Let $\{s\}$ be an optimal solution
for $P(\Phi)$ and assume, by contradiction, that $\Phi$ is
satisfiable. Then there is a set of hypotheses $S\subseteq
H-\{s\}$ that encodes a satisfying truth-value assignment for
$\Phi$ and has cost $0$. However, this contradicts the fact that
the solution $\{s\}$, which has cost $1$, is optimal.
\end{proof}

 We next determine the complexity of deciding
 the relevance of an hypothesis.

\begin{theorem}\label{theo:relevance}
Deciding whether an hypothesis is relevant for a $\PAP$ $\gpst$ is
$\DeltaP{2}$-complete. Hardness holds even if $P$ is a
constraint-free positive program.
\end{theorem}
\begin{proof}(\emph{Membership}).
Let $\gpst$ be a $\PAP$ and let $h\in H$ be a hypothesis.
First we compute the maximum value {\it max} that the
function ${\it sum}_\gamma$ may return over all sets $H'\subseteq H$. Note
that {\it max} is polynomial-time computable from $\gp$,
because $\gamma$ is a polynomial-time computable function.
It follows that its size $|{\it max}|=\log {\it max}$
is $O(|\gp|^k)$, for some constant $k\geq 0$,
because the output size of a polynomial-time computable function is
polynomially-bounded, as well.

Then, by a binary search on $[0, {\it max}]$,
we compute the cost $c$ of the optimal solutions for $\gp$:
at each step of this search, we are given a threshold $s$ and
we call an $\NP$ oracle to know
whether there exists an admissible solution below $s$.
After, $\log {\it max}$ steps at most,
this procedure ends, and we get the value $c$.
Finally, we ask another $\NP$ oracle whether
there exists an admissible solution containing $h$ and whose cost is $c$.
Note that the number of steps and hence the number of oracle calls
is polynomial in the input size,
and thus deciding whether $h$ is relevant is in $\DeltaP{2}$.

\

\noindent (Hardness). We reduce the $\DeltaP{2}$-complete problem
of deciding whether a TSP instance $I$ has a unique optimal tour
\cite{papa-84} to the relevance problem for the $\PAP$ $\gp\ =
\tuple{H,\gamma, P, O}$, defined below, whose optimal solutions
encode, intuitively, pairs of optimal tours. The set of hypotheses
is $H= \{ c(i,j),c'(i,j) \ |\ 1 \leq i,j \leq n\} \cup \{h_{eq},
h_{\it diff}\}$, where $c(i,j)$ (resp., $c'(i,j)$) says that the
salesman visits city $j$ immediately after city $i$, according to
the tour encoded by the atoms with predicate $c$ (resp., $c'$).
Moreover, the special atoms $h_{eq}$ and $h_{\it diff}$ encode the
hypotheses that such a pair of tours represents in fact a unique
optimal tour, or two distinct tours.

For each pair of cities $c_i,c_j$,
the penalty function $\gamma$ encodes the cost function $w$ of travelling
from $c_i$ to $c_j$,
that is, $\gamma(c(i,j)) = w(i,j)$ and $\gamma(c'(i,j)) = w(i,j)$.
Moreover, for the special atoms, define
$\gamma(h_{eq})=1$ and $\gamma(h_{\it diff})=0.5$.

The program $P$, shown below, is similar to the TSP
encoding described in Section \ref{sec:tsp}:%
{
\[
\begin{array}{lrll}
(1) & visited(I) & \derives\   visited(J), c(J,I). &  \\
(2) & visited(1) & \derives\   c(J,1). &  \\
(3) & badTour & \derives\ c(I,J), c(I,K), J \neq K. &\\
(4) & badTour &\derives\  c(J,I), c(K,I), J \neq K. &\\
(1') & visited'(I) & \derives\   visited'(J), c'(J,I). &  \\
(2') & visited'(1) & \derives\   c'(J,1). &  \\
(3') & badTour & \derives\  c'(I,J), c'(I,K), J \neq K. &\\
(4') & badTour &\derives\  c'(J,I), c'(K,I), J \neq K. &\\
(5) & \diff  &\derives\  c(I,J), c'(I,K), J \neq K. &\\
(6) & ok & \derives \ h_{eq}. & \\
(7) & ok & \derives \ h_{\it diff}, \diff. &
\end{array}
\]
}
The observations are $O =\{\ok, not \ badTour \}\cup
\{ \visited(i), \visited'(i) \mid \leq i\leq n\}$.

Note that every admissible solution $S$ for $\gp$ encodes two legal tours
 for $I$, through atoms with predicates $c$ and $c'$.
Moreover, $S$ contains either $h_{eq}$ or $h_{\it diff}$, in order to
 derive the observation $\ok$.
Furthermore, if $S$ is optimal, then at most one of these special
 atoms belongs to $S$, because one is sufficient to get $\ok$.
However, if the chosen atom is $h_{\it diff}$, $\ok$ is derivable only
 if $\diff$ is true, i.e., the two encoded tours are different, by rule (5).

Let $\tmin$ be the cost of an optimal tour of $I$.
Then, the best admissible solution $S$ such that $h_{eq}\in S$
has cost $2 \tmin +1$, because it should contain the hypotheses
encoding two (possibly identical) optimal tours of $I$, and the atom $h_{eq}$.

We show that there is a unique optimal tour for $I$ if and only if
$h_{eq}$ is a relevant hypothesis for $\gp$.

(\emph{Only if}). Let $T$ be the unique optimal tour $T$ for $I$,
and $S$ the admissible solution for $\gp$ such that $h_{eq}\in S$ and
both the atoms with predicate $c$ and those with predicate $c'$ encode
the tour $T$.
Then, $S$ is an optimal solution, because any admissible solution $S'$
that does not contain $h_{eq}$ should contain both $h_{\it diff}$ and
$\diff$. Since $T$ is the unique optimal tour, any other legal tour
$T'$ has cost $\tmin+1$, at least.
Hence, $sum_\gamma(S')\geq \tmin + (\tmin +1) + 0.5 > sum_\gamma(S)$.
Thus, $h_{eq}$ is relevant for $\gp$, because belongs to the optimal
solution $S$.

(\emph{If}). If $h_{eq}$ is relevant for $\gp$,
there is an optimal solution $S$ such that $h_{eq}\in S$.
Recall that $sum_\gamma(S) = 2\tmin + 1$.
Assume by contradiction that there are two distinct
optimal tours $T$ and $T'$ for $I$,
and let $S'$ be an admissible solution such that:
its atoms with predicates $c$ and $c'$ encode the distinct tours $T$ and $T'$,
and both $\diff$ and $h_{\it diff}$ belong to $S'$.
Then, $sum_\gamma(S') = 2 \tmin + 0.5 < sum_\gamma(S)$, a contradiction.

Finally, note that $P$ is constraint-free positive program, and
both $P$ and its ground instantiation can be computed in
polynomial time from the instance $I$.
\end{proof}

Not surprisingly, the necessity problem has the same complexity as the
relevance problem.

\begin{theorem}\label{theo:necessity}
Deciding whether an hypothesis is necessary for a $\PAP$ $\gpst$
is $\DeltaP{2}$-complete. Hardness holds even if $P$ is a
constraint-free positive program.
\end{theorem}
\begin{proof}
(\emph{Membership}).
Let $\gpst$ be a $\PAP$ and let $h\in H$ be a hypothesis.
We compute the cost $c$ of the optimal solutions for $\gp$,
as shown in the proof of Theorem~\ref{theo:relevance}.
Finally, we ask an $\NP$ oracle whether
there exists an admissible solution whose cost is $c$, and does not
contain $h$. If the answer is no, then $h$ is a necessary hypothesis.
Clearly, even in this case, a polynomial number of calls to $\NP$
oracle suffices, and thus the problem is in $\DeltaP{2}$.

\

\noindent (Hardness).
Let $I$ be a TSP instance and $\gp$ the $\PAP$ defined in the proof of
Theorem~\ref{theo:relevance}.
Note that the same reasoning as in the above proof shows that $I$ has a
unique optimal tour if and only if $h_{eq}$ is a necessary hypothesis
for $\gp$.
\end{proof}

\begin{theorem}\label{theo:optimal}
Computing an optimal solution for a $\PAP$ $\gpst$ is
$\FPNP$-complete. Hardness holds even if $P$ is a constraint-free
positive program.
\end{theorem}
\begin{proof}(\emph{Membership}).
Let $M$ be a deterministic Turing transducer $M$ with oracles in $\NP$ that
act as follows.
First, $M$ checks in $\NP$ whether $\gp$ is consistent, as shown in
the proof of Theorem \ref{theo:consistency}. If this not the case,
then $M$ halts and writes on its output tape some special symbol
encoding the fact that $\gp$ is inconsistent. Otherwise,
$M$ computes with a polynomial number of steps
the value $c$ of the optimal
solutions for $\gp$, as shown in the proof of Theorem \ref{theo:relevance}.
Now, consider the following oracle $O$:
given a set of hypotheses $S$, decide whether there is an admissible
solution for $\gp$ whose cost is $c$. It is easy to see that $O$ is in
$\NP$ (we describe a very similar proof in the membership part of the
proof of Theorem \ref{theo:admissible}).

The transducer $M$ maintains in its worktape (the encoding of) a set
of hypotheses $S$, which is initialized with $\emptyset$.
Then, for each hypotheses $h\in H$, $M$ calls the
oracle $O$ with input $S\cup \{h\}$. If the answer is yes, then $M$
writes $h$ on the output tape and adds $h$ to the set $S$.
Otherwise, $S$ is not changed, and $M$ proceeds with the next
candidate hypothesis.
It follows that, after $|H|$ of these steps, the output tape encodes an
optimal solution of $\gp$.

\

\noindent
(\emph{Hardness}).
Immediately follows from our encoding of the TSP problem shown in
Section \ref{sec:tsp}, and the fact
that this problem is $\FPNP$-complete \cite{papa-94}.
\end{proof}

\section{Implementation Issues}\label{sec:implementation}

In this section, we describe the implementation of a system
supporting our formal model of abduction with penalties over logic
programs. The system has been implemented as a front-end for the
\dlv\ system. Our implementation is based on a translation from
such abduction problems to logic programs with weak constraints
\cite{bucc-etal-99b}, that we show to be both sound and complete.
We next describe the architecture of the prototype. We then
briefly recall Logic Programming with Weak Constraints (the target
language of our translation), define precisely our translation
algorithm and prove its correctness.

\subsection{Architecture}

Figure \ref{fig:architecture} shows the architecture of the
new abduction front-end for the \dlv\ system,
which implements the framework of abduction with penalization from
logic programs, and is already incorporated in the current release of
\dlv\ (available at the \dlv\ homepage \url{www.dlvsystem.com}).

\begin{figure}[h]
\vspace{0.5 cm} \epsfxsize=0.9\textwidth
\epsfbox{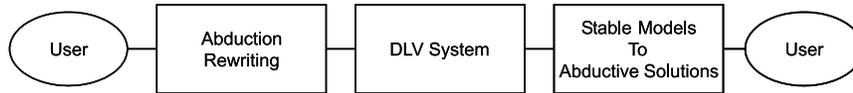} \caption{System
Architecture}\label{fig:architecture}
\end{figure}

A problem of abduction in \dlv\ consists of three separate files
encoding the hypotheses, the observations, and the logic program.
The first two files have extensions {\tt .hyp} and {\tt .obs},
respectively, while no special extension is required for the
logic-program file. The abduction with penalization front-end is
enabled through the option {\tt -FDmincost}. In this case, from
the three files above, the Abduction-Rewriting module builds a
logic program with weak constraints, and run \dlv\ for computing a
best model $M$ of this logic program. Then, the
Stable-Models-to-Abductive-Solutions module extracts an optimal
solution from the model $M$.

For instance, consider the network problem in Example \ref{ex:network},
and assume that the facts encoding the hypotheses are stored in the
file {\tt network.hyp} , the facts encoding the observations are
stored in the file {\tt network.obs}, and the logic program is stored
in the file {\tt netwok.dl}. Then, the user may obtain an optimal
solution for this problem by running:

{\tt dlv -FDmincost network.dl network.hyp network.obs}

By adding option {\tt -wctrace} the system prints also the
(possibly not optimal) solutions that are found during the
computation. This option is useful to provide some solution
to the user as soon as possible.
Note that the ``quality'' of the solutions increases
monotonically (i.e., the cost decreases), and the system gradually
converges to optimal solutions.

Note that the current release deals with integer penalties only; however, it
can be extended easily to real penalties.

\subsection{Logic Programming with Weak Constraints}
\label{sec:dlv}

We first provide an informal description of the \lpw language by
examples, and we then supply a formal definition of the syntax and
semantics of \lpw.


\subsubsection{\lpw by Examples}

Consider the problem SCHEDULING, consisting in the scheduling of
course examinations. We want to assign course exams to time slots
in such a way that no couple of exams are assigned to the same
time slot if the corresponding courses have some student in common
-- we call such courses ``incompatible''. Supposing that there are
three time slots available, $\tt ts_1$, $\tt ts_2$ and $\tt ts_3$,
we express the problem in \lpw by the following program
$\p_{sch}$:

\[
\begin{array}{ll}
r_1:& \  {\tt assign(X,ts_1) \derives \ course(X), not \ assign(X,ts_2), not \ assign(X,ts_3).}\\
r_2:& \  {\tt assign(X,ts_2) \derives \ course(X), not \ assign(X,ts_1), not \ assign(X,ts_3).}\\
r_3:& \ {\tt assign(X,ts_3) \derives \ course(X), not \ assign(X,ts_1), not \ assign(X,ts_2).}\\
s_1:& \ {\tt \derives \ assign(X,S), assign(Y,S), commonStudents(X,Y,N).}\\
\end{array}
\]

Here we assumed that the courses and the pair of courses with
common students are specified by input facts with predicate
$\tt course$ and $\tt commonStudents$, respectively.  In particular,
$\tt commonSudents(a,b,k)$ means that there are $\tt k>0$ students who
should attend both course $\tt a$ and course $\tt b$. Rules $r_1$, $r_2$
and $r_3$ say that each course is assigned to one of the
three time slots $\tt ts_1$, $\tt ts_2$ or $\tt ts_3$; the strong constraint
$s_1$ expresses that no two courses with
some student in common can be assigned to the same time slot. In
general, the presence of strong constraints modifies the semantics
of a program by discarding all models which do not satisfy some of
them. Clearly, it may happen that no model satisfies all
constraints. For instance, in a specific instance of above
problem, there could be no way to assign courses to time slots
without having some overlapping between incompatible courses.  In
this case, the problem does not admit any solution. However, in
real life, one is often satisfied with an approximate solution, in
which constraints are satisfied as much as possible. In this
light, the problem at hand can be restated as follows (APPROX
SCHEDULING): ``assign courses to time slots trying to avoid overlapping
courses having students in common.'' In order to express this problem
we introduce the notion of {\em weak} constraint, as shown by the
following program $P_{a\_sch}$:
\[
\begin{array}{ll}
r_1:&\ {\tt assign(X,ts_1) \derives \ course(X), not \ assign(X,ts_2) not \ assign(X,ts_3).}\\
r_2:&\ {\tt assign(X,ts_2) \derives \ course(X), not \ assign(X,ts_1) not \ assign(X,ts_3).}\\
r_3:&\ {\tt assign(X,ts_3) \derives \ course(X), not \ assign(X,ts_1) not \ assign(X,ts_2).}\\
w_1:&\ {\tt \weakderives \ assign(X,S), assign(Y,S),
commonStudents(X,Y,N).}
\end{array}
\]
 From a syntactical point of view, a weak constraint is like a
strong one where the implication symbol $\tt\derives$ is replaced
by $\tt \weakderives$.  The semantics of weak constraints
minimizes the number of violated instances of constraints.  An
informal reading of the above weak constraint $w_1$ is:
``\underline{preferably}, do not assign the courses $\tt X$ and
$\tt Y$ to the same time slot if they are incompatible''. Note
that the above two programs $\p_{sch}$ and $\p_{a\_sch}$ have
exactly the same preferred models if all incompatible courses can
be assigned to different time slots (i.e., if the problem admits
an ``exact'' solution).

In general, the informal meaning of a weak constraint, say,
${\tt \weakderives B.}$, is ``try to falsify $\tt B$'' or ``$\tt B$ is preferably
false'', etc. Weak constraints are very powerful for capturing the
concept of ``preference'' in commonsense reasoning.

Since preferences may have, in real life, different
``importance'', weak constraints in \lpw can be supplied with
different weights, as well.\footnote{Note that weights are
meaningless for strong constraints, since all of them {\em must}
be satisfied.}  For instance, consider the course scheduling
problem: if overlapping is unavoidable, it would be useful to
schedule courses by trying to reduce the overlapping ``as much as
possible'', i.e. the number of students having some courses in
common should be minimized. We can formally represent this problem
(SCHEDULING WITH WEIGHTS) by the following program $P_{w\_sch}$:
\[
\begin{array}{ll}
r_1:&\ {\tt assign(X,ts_1) \derives \ course(X), not \ assign(X,ts_2) not \ assign(X,ts_3).}\\
r_2:&\ {\tt assign(X,ts_2) \derives \ course(X), not \ assign(X,ts_1) not \ assign(X,ts_3).}\\
r_3:&\ {\tt assign(X,ts_3) \derives \ course(X), not \ assign(X,ts_1) not \ assign(X,ts_2).}\\
w_1:&\ {\tt \weakderives \ assign(X,S), assign(Y,S), commonStudents(X,Y,N). }\ \ \ {\tt [N:]} \\
\end{array}
\]

The preferred models (called {\em best models}) of the above
program are the assignments of courses to time slots that minimize
the total number of ``lost'' lectures.


\subsubsection{Syntax and Semantics}

A \textit{weak constraint} has the form
\begin{dlvcode}
\weakderives
L_1,\cdots, L_m. \ \ [w:]
\end{dlvcode}

where each $\tt L_i$, $\tt{1 \leq \ i \leq m}$, is a literal and $\tt w$
is a term that represents the \textit{weight}.%
\footnote{In their general form, weak constraints are labelled by pairs
 $\tt[w:\ell]$, where $\tt w$ is a weight and $\tt \ell$ is a priority level.
 However, in this paper we are not interested in priorities and we thus
 describe a simplified setting, where we only deal with weights.}
In a ground (or instantiated) weak constraint, $\tt w$ is a nonnegative integer.
If the weight $\tt w$ is omitted, then its value is 1, by default.

An $\lpw$ program $P$ is a finite set of rules and constraints (strong and weak).
If $P$ does not contain weak constraints, it is called a
{\em normal} logic program.

Informally, the semantics of an $\lpw$ program $P$ is given by the
stable models of the set of the rules of $P$ satisfying all
strong constraints and minimizing the sum of weights of violated
weak constraints.

Let $R$, $S$, and $W$ be the set of ground
instances of rules, strong constraints, and weak constraints of an
$\lpw$ program $P$, respectively.
A {\em candidate model} of $P$ is a stable model of $R$ which
satisfies all strong constraints in $S$.
A weak constraint $c$ is satisfied in $I$ if
some literal of $c$ is false w.r.t. $I$.

We are interested in those
candidate models that minimize the sum of weights of violated
weak constraints. More precisely, given a candidate model $M$
and a program $P$, we introduce an objective function $\HP(M)$,
defined as:

\[ \HP(M) = \sum_{c \in {\it Violated}_M^P} {\it weight}(c) \]

\noindent where ${\it Violated}_M^P$ = \{ $c \in W \mid c$ is a weak
constraint violated by $M$ \} and ${\it weight}(c)$ denotes the weight of
the weak constraint $c$.
A candidate model $M$ of P is a \textit{best model} of P if
$\HP(M)$ is the minimum over all candidate models of P.

As an example, consider the following program $P_s$:

\begin{displaymath}\begin{array}{l@{\hspace{1cm}}l}
{\tt a \derives \ c, not \ b.} & {\tt \weakderives \ a,c.} \ \ {\tt [1:] } \\
{\tt c. } & { \tt \weakderives \ b.  } \ \ {\tt [2:]}\\
{\tt b \derives \ c, not \ a.} & {\tt \weakderives \ a. }\ \ {\tt [1:]} \\
& {\tt \weakderives \ b,c.} \ \ {\tt [1:]} \\
\end{array}\end{displaymath}

\noindent The stable models for the set ${ \tt \{\ c.\  \  \ a
\derives \ c, not \  b.\  \  \ b \derives  \ c, not \ a.\} }$ of
ground rules of this example are $\HPs(M_1)=\mathtt{\{a,c\}}$ and
$\HPs(M_2)=\mathtt{\{b,c\}}$, they are also the candidate models,
since there is no strong constraint. In this case, $\HPs(M_1)=2$,
and $\HPs(M_2)=3$. So $M_1$ is preferred over $M_2$ ($M_1$ is a
best model of $\p_s$).

\subsection{From Abduction with Penalization to Logic
Programming with Weak Constraints}\label{sec:architecture}
\begin{figure}
\begin{tabbing}
{\bf Input:} A $\PAP$ $\gp$=$\tuple{H,P,O,\gamma}$. \\
{\bf Output:} A logic program with weak constraints $\lpwp$. \\
\\
\bFunction\ AbductionTo$\lpw(\gp : \PAP)$ : \lpw\\
\bvar\ \ \=$i$, $j$: $Integer$;\\
         \>$\gq$ : \lpw;\\
\bbegin \=\\
(1)     \>$\gq$:= $P$;\\
(2)     \>\mbox{Let $H = \tuple{h_1,\ldots,h_n}$;}\\
(3)     \>\bfor\ \=$i:=1$ {\bf to} $n$ {\bf do}\\
(4)     \>       \> add to \= $ \gq$ the following three clauses\\
        \>  (4.a)\>       \>${\tt \lrule{h_i}{\_sol(i)}}$\\
        \>  (4.b)\>       \>${\tt \lrule{\_sol(i)}{not \ \_nsol(i)}}$ \\
        \>       \>       \>${\tt \lrule{\_nsol(i)}{not \ \_sol(i)}}$ \\
        \>  (4.c)\>     \>${\tt \weakderives h_i  \  \ \ [\gamma(h_i):]}$.\\
(5)     \>\bendfor\\
(6)     \>\mbox{Let $O = \{ o_1,\ldots ,o_m\}$;}\\
(7)     \>\bfor\ \=$j:=1$ \bto\ $m$ \bdo\\
(8)     \>       \>\bif\ \=$o_j$ \mbox{is a positive literal ``$a$''}\\
(9)    \>       \>      \>\bthen\ add to $\gq$ the constraint ${\tt \lrule{}{\naf a}}$\\
(10)    \>       \>      \>\belse\ \=$(*\ o_j$ \mbox{is a negative literal ``$\naf a$''}\ $*)$\\
(11)    \>       \>      \>        \>add to $\gq$ the constraint ${\tt \lrule{}{a}}$\\
(12)    \>\bendfor\\
(13)    \>\breturn\ $\gq$;\\
\bend
\end{tabbing}
\caption{Translating a $\PAP$ $\gp$ into a logic program $\lpwp$}
\label{alg:abduction}
\end{figure}

\begin{figure}
\begin{tabbing}
{\bf Input:} \=A stable model $M$ of $\lpwp$, where $\gp$ is $\tuple{H,P,O,\gamma}$. \\
{\bf Output:} A solution of $\gp$.\\
\\
\bFunction\ ModelToAbductiveSolution($M : {\it AtomsSet}$): {\it
AtomsSet}\\
\bvar\ $S$ : {\it AtomsSet};\\
\bbegin\ \=\\
      \>\breturn\ $H \cap M$;\\
\bend
\end{tabbing}

\caption{Extracting a solution of $\gp$ from a stable model of
$\lpwp$} \label{algorithm-model-diagnoses}
\end{figure}


Our implementation of abduction from logic programs with penalization
is based on the algorithm shown in Figure \ref{alg:abduction},
which transforms a $\PAP$
$\gp$ into a logic program $\lpwp$ whose stable models correspond
one-to-one to abductive solutions of $\gp$.

We illustrate this algorithm by an example.
\begin{example}\label{translation}
Consider again the {\em Network Diagnosis} problem described in
Example \ref{ex:network}. The translation algorithm constructs an
$\lpw$ program $\gq$. First, $\gq$ is initialized with the logic
program $P$. Therefore, after Step 1, $\gq$ consists of the set of
facts encoding the network and of the following rules:
\[\small
\begin{array}{l@{\ \derives \ }l}
\tt reaches(X,X) &  \tt node(X), not \ \; \tt offline(X). \\
\tt    reaches(X,Z) & \tt reaches(X,Y), \  \tt connected(Y,Z),\
not \ \; {\tt offline}(Z).
\end{array}
\]
Then, in the loop 3-5, the following groups of
rules and weak constraints are added to $\gq$.
\begin{itemize}
\item[ At Step 4.a:]
\[\small
\begin{array}{@{\hspace{-0.4cm}}l@{\hspace{0.4cm}}l@{\hspace{0.4cm}}l@{\hspace{0.4cm}}l}
\tt offline(a) \derives \tt \_sol(1).  &  \tt offline(b) \derives \tt \_sol(2). & \cdots &\tt offline(f) \derives \tt \_sol(6).
\end{array}
\]
\item[At Step 4.b:]
\[\small
\begin{array}{@{\hspace{-0.4cm}}l@{\hspace{0.4cm}}l@{\hspace{0.4cm}}l@{\hspace{0.4cm}}l}
\tt \_sol(1) \derives \tt not \ \_nsol(1). & \tt \_sol(2) \derives
not \ \tt \_nsol(2). & \cdots & \tt \_sol(6) \derives not \ \tt \_nsol(6).\\
\tt \_nsol(1) \derives not \ \tt \_sol(1). & \tt
\_nsol(2) \derives not \ \tt \_sol(2). & \cdots & \tt \_nsol(6) \derives not \ \tt \_sol(6).
\end{array}
\]
\item[At Step 4.c:]

\[\small
\begin{array}{ll}
 \tt \weakderives offline(a). & \tt [\gamma({\tt offline(a)}):] \\
 \tt \weakderives offline(b). & \tt [\gamma({\tt offline(b)}):] \\
 \cdots & \\
\tt \weakderives offline(f). & \tt [\gamma({\tt offline(f)}):]
\end{array}
\]
\end{itemize}

The above rules select a set of hypotheses as a candidate solution,
and the weak constraints are weighted according to the hypotheses penalties.
Thus, weak constraints allow us to compute
the abductive solutions minimizing the sum of the hypotheses
penalties, that is, the optimal solutions.

Finally, to take into account the observations,
the following constraints are added to $\gq$
in the loop 7-12:

\[\small
\begin{array}{l@{\ \derives \ }l}
 & \tt not \ offline(a). \\
 & \tt not \ offline(e). \\
 & \tt not \ reaches(a,e). \\
\end{array}
\]

This group of (strong) constraints is added to $\gq$ in order to
discard stable models that do not entail the observations.

Note that, since in this example all observations are positive literals,
Step 11 is never executed.
\end{example}

The logic program $\lpwp$ computed by this algorithm is then evaluated by the
{\dlv} kernel, which computes its stable models. For each model $M$
found by the kernel, the ModelToAbductiveSolution function (shown in
Figure \ref{algorithm-model-diagnoses}) is called in
order to extract the abductive solution corresponding to $M$.

The next theorem states that our strategy is
sound and complete. For the sake of presentation, its proof
is reported in Appendix \ref{app:proofs}.

\begin{theorem}\label{theo:algo}
\noindent
(\emph{Soundness})
For each best model $M$ of $\lpwp$, there exists an optimal
solution $A$ for $\gp$ such that $M \cap H = A$.

\noindent
(\emph{Completeness})
For each optimal solution $A$ of $\gp$, there exists a best
model $M$ of $\lpwp$ such that $M \cap H = A$.
\end{theorem}

\section{Related Work}\label{sec:related}

Our work is evidently related to previous studies on semantic and
knowledge representation aspects of abduction over logic programs
\cite{kaka-manc-90b,lifs-turn-94b,kaka-etal-00,dene-desc-98,lin-you-02},
that faced the main issues concerning this form of non-monotonic reasoning,
including detailed discussions on how such a formalism may be used
effectively for knowledge representation -- for a nice survey,
see \cite{dene-kaka-02}.

However, all these works concerning abduction from logic programs do not deal
with penalties.
The present paper focuses on this kind of abductive reasoning from logic
programs, and our computational complexity analysis extends
and complements the previous studies on the complexity of abductive
reasoning tasks \cite{eite-gott-95,eite-etal-97k}.

The optimality criterion we use in this paper for identifying the
best solutions (or explanations) is the minimization of the sum of
the penalties associated to the chosen hypotheses. Note that this
is not the only way of preferring some abductive solutions over
others. In fact, the traditional approach, also considered in the
above mentioned papers, is to look for minimal solutions
(according to standard set-containment). From our complexity
results and from the results presented in \cite{eite-etal-97k}, it
follows that the (set) minimal explanation criterion is more
expensive than the one based on penalties, from the computational
point of view. Moreover, this kind of weighted preferences has
been recognized as a very important feature in real applications.
Indeed, in many cases where quantitative information plays an
important role, using penalties can be more natural than using
plain atoms and then studying some clever program such that
minimal solutions correspond to the intended best solutions. As a
counterpart, if necessary, in the minimal-explanations framework
we can represent some problems belonging to high complexity
classes that cannot be represented in the penalties framework. It
follows that the two approaches are not comparable, and the choice
should depend on the kind of problem we have to solve.

Another possible variation concerns the semantics for logic
programs, which should not be necessarily  the stable model
semantics. For instance, in \cite{pere-etal-91}, a form of
hypothetical reasoning is based on the well-founded semantics. In
some proposals, the semantics is naturally associated to a
particular optimality criterion, as for \cite{inou-saka-99}, where
the authors consider prioritized programs under the preferred
answer set semantics.

A similar optimization criterion is proposed for the logic programs with
consistency-restoring rules (cr-rules) described in \cite{bald-gelf-03}.
Such rules may contain preferences and are used for making a give program consistent,
if no answer set can be found. Firing some of these rules and hence deriving some
atoms from their heads corresponds in some way to the hypotheses selection in
abductive frameworks. Indeed, the semantics of this language is based on a transformation
of the given program with cr-rules into abductive programs.

Such optimization criteria induce partial orders among solutions,
while we have a total order, determined by the sum of penalties.
We always have the minimum cost and the solutions with this cost constitute
the equivalence class of optimal solutions.
Note that even these frameworks are incomparable with our approach based on penalties,
and which approach is better just depends on the application one is interested in.

Since we provide also an implementation of the proposed framework,
our paper is also related to previous work on abductive logic programming
systems \cite{nuff-kaka-01,kaka-etal-01}.
More links to systems and to some interesting applications of
abduction-based frameworks to real-world problems can be found at
the web page \cite{toni-web-03}.

We remark that we are not proposing an algorithm for solving
optimizations problems. Rather, our approach is very general and
aims at the representation of problems, even of optimization
problems, in an easy and natural way through the combination of
abduction, logic programming, and penalties. It is worthwhile
noting that our rewriting procedure into logic programs with weak
constraints (or similar kind of logic programs) is just a way for
having a ready-to-use implementation of our language, by
exploiting existing systems, such as $\dlv$
\cite{eite-etal-98a,leon-etal-2002-dlv} or {\it smodels}
\cite{niem-simo-97,simo-etal-2002}. Differently, operations
research is completely focused on finding solutions to
optimization problems, regardless of representational issue. In
this respect, it is worthwhile noting that, in principle, one can
also use techniques borrowed from operations research for
computing our abductive solutions (e.g., by using integer
programming).

A second point is that in the operations research field one can find
algorithms specifically designed for solving, e.g., only TSP instances,
or even only some particular TSP instances \cite{tsp-book-02}.
It follows that our general approach is not in competition with operations
research algorithms. Rather, such techniques can be exploited profitably
for computing abductive solutions, if we know that the programs under consideration
are used for representing some restricted class of problems.

\section{Conclusion}\label{sec:conclusion}

We have defined a formal model for abduction with
penalization from logic programs.
We have shown that the proposed
formalism is highly expressive and it allows to encode relevant
problems in an elegant and natural way. We have carefully analyzed
the computational complexity of the main problems
arising in this abductive framework.
The complexity analysis shows an
interesting property of the formalism: ``negation comes for
free'' in most cases, that is, the addition of negation does
not cause any further increase to the complexity of the abductive
reasoning tasks (which is the same as for positive programs).
Consequently, the user can enjoy the knowledge representation power of
nonmonotonic negation without paying high costs in terms of
computational overhead.

We have also implemented the proposed language on top of the \dlv\
system. The implemented system is already included in the current
\dlv\ distribution, and can be freely retrieved from \dlv\
homepage \url{www.dlvsystem.com} for experiments.

It is worthwhile noting that our system is not intended to be a
specialized tool for solving optimization problems. Rather, it is
to be seen as a general system for solving knowledge-based
problems in a fully declarative way. The main strength of the
system is its high-level language, which, by combining logic
programming with the power of cost-based abduction, allows us to
encode many knowledge-based problems in a simple and natural way.
Evidently, our system cannot compete with special purpose algorithms
for, e.g., the Travelling Salesman Problem; but it could be used
for experimenting with nonmonotonic declarative languages.
Preliminary results of experiments on the {\em Travelling Salesman
Problem} and on the {\em Strategic Companies Problem} (see Section
\ref{sec:kr}) show that the system can solve also instances
of a practical interest (with more than 100 companies for
Strategic Companies and 30 cities for Travelling Salesman).

\section*{Acknowledgments}

The authors are grateful to Giovambattista Ianni for some useful
discussions on the system implementation.

This work was supported by the European Commission under project
 IST-2002-33570 INFOMIX, IST-2001-32429 ICONS, and IST-2001-37004 WASP.

\newcommand{\SortNoOp}[1]{}

\appendix
\section{Proof of Theorem \ref{theo:algo}}\label{app:proofs}

In this appendix, we prove that the rewriting approach described in
Section \ref{sec:implementation} is sound and complete.

First, we recall an important result on
the {\em modularity} property of logic programs under the
stable model semantics, proved in \cite{eite-etal-97f}.

Let $P_1$ and $P_2$ be two logic programs.
We say that $P_2$ {\em potentially
\ uses} $P_1$ ($P_2 \rhd P_1$) iff each predicate that occurs in
some rule head of $P_2$ does not occur in $P_1$.

Moreover, given a set of atoms $M$ and a program $P$, we denote by $M|_P$
the set of all atoms from $M$ that occur in $P$, i.e.,
$M|_P = M\cap B _{P}$.

\begin{proposition}\cite{eite-etal-97f} \label{theo:modularity}
Let $P = P_1 \cup P_2$ be a logic program such that $P_2$
potentially uses $P_1$. Then,

(i) for every $M \in \SM(P)$, $M|_{P_1} \in \SM(P_1)$;

(ii) $\SM(P)= \bigcup_{M \in \SM(P_1)} \SM(P_2 \cup\facts(M))$.
\end{proposition}

\begin{lemma} \label{lem:modul}
Let $P = P_1 \cup P_2$ be a logic program such that $P = P_1 \cup
P_2$ and $P_2 \rhd P_1$. Then, for every $M\in\SM(P)$,
$M|_{P_2}$ is a stable model for
$P_2 \cup \facts(M|_{P_1} \cap B_{P_2})$.
\end{lemma}
\begin{proof}
 From Proposition~\ref{theo:modularity}~(ii), it follows that there exists
$M_1 \in \SM(P_1)$ such that $M \in \SM(P_2\cup\facts(M_1))$.
We claim that $M_1 = M|_{P_1}$.

\noindent
($M_1 \subseteq M|_{P_1}$).
Immediately follows from the fact that
$M_1 \subseteq M$, because $M\in \SM(P_2\cup\facts(M_1))$.

\noindent
($M|_{P_1}\subseteq M_1$). Suppose by contradiction
that there exists an atom $a\in M$
such that $a \in M|_{P_1}$ but $a \notin M_1$.
It follows that $a$ is not defined in $P_1$ and thus there exists some rule
$r$ of $P_2$ having $a$ in its head. However, this is impossible,
as we assumed that $P_2 \rhd P_1$. Contradiction.

Thus, $M$ is a stable model of $P_2\cup\facts(M|_{P_1})$.
Let $A = M|_{P_1}\cap B_{P_2}$ and $X = M|_{P_1} - A$;
whence, $\facts(M|_{P_1})= \facts(A)\cup\facts(X)$.
Note that $X$ contains all and only the atoms of $M$ not occurring in $P_2$.
Therefore, it is easy to see that
$M-X$ is a stable model for $P_2\cup(\facts(M|_{P_1})-\facts(X))$,
which is equal to $P_2\cup\facts(A)$.
Moreover, observe that $M-X = M|_{P_2}$, and
thus we get $M|_{P_2} \in \SM(P_2\cup\facts(A))$.
\end{proof}

For the sake of presentation, we assume hereafter a given $\PAP$
problem $\gpst$ is fixed, and let $\lpwp= P\cup P_{hyp}\cup
P_{obs}$ be the program computed by the function
AbductionTo$\lpw(\gp)$, where $P_{hyp}$ is the set of rules and
weak constraints obtained by applying steps (3)-(5), and $P_{obs}$
is the set of strong constraints obtained by applying steps
(7)-(12).

\begin{lemma}\label{lemma:stable-adm}
For each stable model $M$ of $\lpwp$,
\begin{itemize}
\item[(a)]
there exists an admissible solution $A$ for $\gp$ such that $M\cap H =A$, and
\item[(b)]
$sum_\gamma(A) = \HP(M)$.
\end{itemize}
\end{lemma}
\begin{proof}
\emph{(Part a).} To show that $M \cap H$ is an admissible solution
for $\gp$ we have to prove that there exists a stable model $M'$
of $P \cup \facts(M\cap H)$ such that, $\forall o \in O$, $o$ is
true w.r.t $M'$.

Let $M'= M|_{P}$. Note that $M'$ is the set of literals obtained
from $M$ by eliminating all the literals with predicate symbol
$\_sol$ and $\_nsol$, i.e. $M'$ is the set of literals without all
atoms which were introduced by the translation algorithm.

Note that $P$ potentially uses $P_{hyp}$. Thus, from
Lemma~\ref{lem:modul}, $M|_{P}$ is a stable model for
$P\cup\facts(C)$, where $C= M|_{P_{hyp}} \cap B_{P}$ and hence $C=
M\cap H$, because only hypothesis atoms from $P_{hyp}$ occur in
$B_{P}$.

Finally, observe that each observation in $O$ is true w.r.t $M'$.
Indeed, since $M$ is a stable model for $\lpwp$, all the
constraints contained in $P|_{obs}$ must be satisfied by $M$.
Moreover, $M$ and $M'$ coincide on all atoms occurring in these
constraints. Thus, all constraints contained in $P_{obs}$ are
satisfied by $M'$, too.

\noindent
\emph{(Part b).} By construction of
$\lpwp$, all weak constraints occurring in this program
involve hypotheses of $\gp$.
In particular, observe that any weak constraint
$\weakderives h  \  \ \ [\gamma(h):]$ is violated by $M$ iff $h$
belongs to $M$. Since $A= M\cap H$, $h$ belongs to $A$, as well,
and its penalty is equal to the weight of the weak
constraint.
It follows that $sum_\gamma(A) = \HP(M)$.
\end{proof}

\begin{lemma}\label{lemma:adm-stable}
For each admissible solution $A$ of $\gp$,
\begin{itemize}
\item[(a)]
there exists a stable model $M$ of $\lpwp$ such that $M \cap H =A$, and
\item[(b)]
$\HP(M)=sum_\gamma(A)$.
\end{itemize}
\end{lemma}
\begin{proof}
\noindent \emph{(Part a).} By Definition~\ref{def:abd-sol}, there
exists a stable model $M' = M'' \cup A$ of $P \cup \facts(A)$,
where $M''\cap A = \emptyset$, such that, $\forall o\in O$, $o$ is
true w.r.t. $M$.

Let $M_H=\{\_sol(i) \mid h_i \in A\}
   \cup \{\_nsol(j) \mid h_j \notin A\}$.

Moreover, let $P'=P \cup \facts(M_H\cup A)$. Note that $P'$ can
also be written as the union of the programs $P \cup \facts(A)$
and $\facts(M_H)$. Since these two programs are completely
disjoint, i.e., the intersection of their Herbrand bases is the
empty set, then the union of their stable models $M'$ and $M_H$,
say $M$, is a stable model of $P'$.

Now, consider the program $P\cup P_{hyp}$, and observe that $P
\rhd P_{hyp}$, and that $M_H \cup A$ is a stable model for
$P_{hyp}$. Then, by Proposition~\ref{theo:modularity}, any stable
model of the program $P'$ is a stable model of $P \cup P_{hyp}$.
Thus, in particular, $M = M' \cup M_H$ is a stable model of $P
\cup P_{hyp}$.

Moreover, it is easy to see that all constraints in $P_{obs}$ are satisfied
by $M$, and thus $M\in \SM(\lpwp)$, too.
Finally, $M$ can be written as $M''\cup A \cup M_H$, and hence
$M\cap H = A$ holds, by definitions of $M''$ and $M_H$.

\noindent
\emph{(Part b).}
Let $h$ be any hypothesis belonging to $A$ and hence contributing
to the cost of this solution.
Note that $\lpwp$ contains the weak constraint
$\weakderives h  \  \ \ [\gamma(h):]$, weighted by $\gamma(h)$
and violated by $M$, as $h\in M$.
It follows that $\HP(M)=sum_\gamma(A)$.
\end{proof}


\vspace{2mm}

\noindent
\emph{Theorem \ref{theo:algo}}\\
(\emph{Soundness})
For each best model $M$ of $\lpwp$, there exists an optimal
solution $A$ for $\gp$ such that $M \cap H = A$.

\noindent
(\emph{Completeness})
For each optimal solution $A$ of $\gp$, there exists a best
model $M$ of $\lpwp$ such that $M \cap H = A$.

\begin{proof}
\emph{(Soundness).} Let $M$ be a best model of $\lpwp$.
 From Lemma \ref{lemma:stable-adm}, $A = M \cap H$ is an admissible
solution for $\gp$, and $sum_\gamma(A) = \HP(M)$.
It remains to show that $A$ is optimal.

By contradiction, assume that $A$ is not optimal.
Then, there exists an admissible solution $A'$ for $\gp$ such that
$sum_\gamma(A') < sum_\gamma(A)$.
By virtue of Lemma \ref{lemma:adm-stable},
we have that there exists a stable model
$M'$ for $\lpwp$ such that $M'\cap H = A'$ and $\HP(M')= sum_\gamma(A')$.
However, this contradicts the hypothesis that $M$ is a best model for $\lpwp$.

\emph{(Completeness).}
Let $A$ be an optimal solution for $\gp$.
By virtue of Lemma \ref{lemma:adm-stable},
 there exists a stable model $M$ for $\lpwp$ such that
 $M \cap H = A$ and $\HP(M)= sum_\gamma(A)$.
 We have to show that $M$ is a best model.

 Assume that $M$ is not a best model. Then,
 there exists a stable model $M'$ for $\lpwp$ such that $\HP(M') < \HP(M)$.
By Lemma \ref{lemma:stable-adm}, there exists an
admissible solution $A'$ for $\gp$ such that $M' \cap H = A'$
and $sum_\gamma(A')= \HP(M')$.
However, this contradicts the
hypothesis that $A$ is an optimal solution for $\gp$.
\end{proof}

\section{A Logic Program for the Travelling Salesman Problem}
\label{app:tsp}

In this section we describe  how to represent the Travelling
Salesman Problem in logic programming.

Suppose that the cities are encoded by a set
of atoms $\{ \tt city(i) \mid 1\leq \tt i\leq n\}$ and that the
intercity traveling costs are stored in a relation $\tt C(i,j,v)$
where $v= w(i,j)$. In abuse of notation, we simply refer to the
number $n$ of cities (which is provided by an input relation) by
itself.

The following program $\pi_1$ computes legal tours and
their costs in its stable models:

{
\[
\begin{array}{@{\hspace*{-3cm}}l@{\hspace*{0.5cm}}r@{\ \derives \ }l@{\hspace{-3cm}}l}
(1) & \tt T(I,J) \vee \tt \tilde{T}(I,J) & \tt c(I,J,\_). \\
(2) &  & \tt T(I,J), T(I,K), J\neq K.  \\
(3) &  & \tt T(I,K), T(J,K), I\neq J.  \\
(4) & \tt visited(1) &\tt T(J,1). \\
(5) & \tt visited(I) & \tt T(J,I), visited(J). \\
(6) &     & \tt  not \ visited(I), city(I). \\
(7) & \tt P\_Value(1,X) & \tt T(1,J), C(1,J,X). \\
(8) & \tt P\_Value(K,X) &  \tt P\_Value(K$-$1,\tt Y), T(K,I), C(K,I,Z), X=Y+Z.\\
(9) & \tt Cost(x) & \tt P\_Value(n,x).
\end{array}
\]
}

The first claus guesses a tour, where $\tt T(I,J)$ intuitively
means that the $I$-th stop of the tour is city $J$ and $\tt
\tilde{T}(I,J)$ that it's not. By the minimality of a stable
model, exactly one of $\tt T(I,J)$ and $\tt \tilde{T}(I,J)$ is
true in it, for each $I$ and $J$ such that $1 \leq I,J \leq n$; in
all other cases, both are false.

The subsequent clauses
    (2)--(6) check that the guess is proper: each stop has attached
at most one city, each city can be attached to at most one stop,
and every stop must have attached some city.
 The rules (7)--(9) compute the cost of the chosen tour, which is given by
 the (unique) atom $\tt Cost(X)$ contained in the model.

 It holds that the stable models of $\pi_1$ correspond one-to-one to the
legal tours.

To reach our goal, we have to eliminate from them those which do
not correspond to optimal tours.  That is, we have to eliminate
all tours $T$ such that there exists a tour $T'$ which has lower
cost. This is performed by a logic program, which basically
tests all choices for a tour $T'$ and rules out each choice that
is not a cheaper tour, which is indicated by a propositional atom
$\tt NotCheaper$. The following program, which is similar to $\pi_1$,
generates all possible choices for $T'$:

\newcommand{\px}[1]{\makebox[4cm][r]{#1}}
\begin{tabbing}
\hspace*{0.3cm} \= (9') \=  \hspace*{4cm} \= \kill \+ \\
$(1') $ \>  \px{ $\tt T'(I,J) \vee \tilde{T}'(I,J)$} \> $\tt \derives c(I,J,\_)$. \\
$(2') $ \>  \px{ $ \tt NotCheaper $} \> ${\tt \derives  T'(I,J), T'(I,K), J\neq K. } $ \\
$(3') $ \>  \px{ $\tt NotCheaper $} \> ${\tt \derives  T'(I,K), T'(J,K), I\neq J. } $ \\
$(4') $ \>  \px{ $\tt NotChosen\_Stop(I,1)   $} \> ${\tt \derives  \tilde{T}'(I,1).} $ \\
$(5') $ \>  \px{ $\tt NotChosen\_Stop(I,J) $} \> ${\tt \derives  \tilde{T}'(I,J),NotChosen\_Stop(I,J}$ - $\tt1). $ \\
$(6') $ \>  \px{ $\tt NotCheaper $} \> ${\tt \derives  NotChosen\_Stop(I,n).} $ \\
$(7') $ \>  \px{ $\tt cnt(1,1) $} \> . \\
$(8') $ \>  \px{ $\tt cnt(K+1,J) $} \> ${\tt \derives  cnt(K,I), {T}'(I,J), J \neq 1.} $ \\
$(9') $ \>  \px{ $\tt NotCheaper$} \> ${\tt \derives  cnt(K,I), {T}'(I,1), K \neq n.} $ \\
$(10') $ \>  \px{ $\tt P\_Value'(1,X) $} \> ${\tt \derives  T'(1,J), C(1,J,X). }$ \\
$(11') $ \>  \px{ $\tt P\_Value'(K,X) $} \> ${\tt \derives   P\_Value'(K}$ - ${\tt 1,Y), T'(K,I), c(K,I,Z), X=Y+Z.}$ \\
$(12') $ \>  \px{ $\tt Cost'(X) $} \> ${\tt \derives  P\_Value'(n,X).}$  \-
\end{tabbing}

The predicates $\tt T'$, $\tt \tilde{T}'$, $\tt P\_Value'$ and $\tt Cost'$ have
the r\^{o}le of the predicates $\tt T$, $\tt \tilde{T}$, $\tt P\_Value$ and
$\tt Cost$ in $\pi_1$. Since we do not allow negation, the test that
for each stop a city has been chosen (rules $(4)$--$(6)$ in
$\pi_1$) has to be implemented
        differently (rules $(4')$--$(9')$~). $\tt NotChosen\_Stop(I,J)$ tells
whether no city $\leq J$ has been chosen for stop $I$. Thus, if
$\tt NotChosen\_Stop(I,n)$ is true, then no city has been chosen for
stop $I$, and the choice for $\tt T'$ does not correspond
        to a legal tour.

The minimal models of $(1')$--$(12')$ which do not contain
$\tt NotCheaper$ correspond one-to-one to all legal tours. By adding
the following rule, each of them is eliminated which does not have
smaller cost than the tour given by $\tt T$: {
\[
\begin{array}{l@{\hspace*{0.5cm}}r@{\ \tt \derives \ }l}
  (13')  &   \tt NotCheaper & \tt Cost(X), Cost'(Y), X \leq Y.
\end{array}
\]
} Thus, if for a legal tour $\tt T$, each choice for $\tt T'$ leads to the
        derivation of $\tt NotCheaper$, then $\tt T$ is an optimal tour.

For the desired program, we add the following rules:
{
\[
\begin{array}{l@{\hspace*{0.5cm}}r@{\ \tt \derives \ }l}
  (14') &  & \tt not \ NotCheaper.\\
  (15') & \tt  P(X_1,\ldots,X_n) &  \tt NotCheaper.\quad,
\end{array}
\]
} for any predicate $\tt P$ that occurs in a rule head of
$(1')$--$(12')$ except $\tt NotCheaper$.  The first rule enforces that
$\tt NotCheaper$ must be contained in the stable model; consequently,
it must be derivable. The other rules derive the maximal extension
for each predicate $\tt P$ if $\tt NotCheaper$ is true, which is a trivial
model for $(1')$--$(12')$. In fact, it is for some given tour $\tt T$
the only model if no choice for $\tt T'$ leads to a tour with cost
smaller than the cost of $\tt T$; otherwise, there exists another
model, which does not contain $\tt NotCheaper$.

Let $\pi_2$ be the program consisting of the rules
$(1')$--$(15')$. Then, it holds that the stable models of $\pi =
\pi_1 \cup \pi_2$ on any instance of TSP
correspond to the optimal tours.%
\footnote{Here, we suppose that the provided universe $U$ of the
database storing the instance is
 sufficiently large for computing the tour values.}
In particular, the optimal cost value, described by $\tt Cost(X)$,
is contained in each stable model. Thus, the program $\pi$
computes on any instance of TSP under the possibility (as well as
certainty) stable model
    semantics in $\tt Cost$ the cost of an optimal tour.

\end{document}